\documentclass[10pt,twocolumn,letterpaper]{article}
\usepackage{amsmath,amsfonts}
\usepackage{array}
\usepackage[caption=false,font=normalsize,labelfont=sf,textfont=sf]{subfig}
\usepackage{textcomp}
\usepackage{stfloats}
\usepackage{verbatim}
\usepackage{graphicx}
\usepackage[accsupp]{axessibility}  

\usepackage{cvpr}              
\usepackage[hyphens]{url}

\definecolor{cvprblue}{rgb}{0.21,0.49,0.74}
\usepackage[pagebackref,breaklinks,colorlinks,allcolors=cvprblue]{hyperref}
\usepackage[capitalize]{cleveref}

\Crefname{table}{Table}{Tables}
\crefname{algorithm}{Algorithm}{Algorithms}
\crefname{figure}{Figure}{Figures}
\crefname{theorem}{Theorem}{Theorems}
\crefname{lemma}{Lemma}{Lemmas}
\crefname{assumption}{Assumption}{Assumptions}
\crefname{definition}{Definition}{Definitions}
\crefname{section}{Section}{Sections}
\Crefname{section}{Appendix}{Appendices}

\usepackage{nicefrac}       
\usepackage{microtype}      
\usepackage{xcolor}         
\usepackage{graphicx}
\usepackage{wrapfig,lipsum,booktabs}
\usepackage{multirow}
\usepackage{amsthm}

\usepackage{bm}
\usepackage{algorithm}
\usepackage{algpseudocode}
\algnewcommand{\algorithmicinitialization}{\textbf{Initialization:}}
\newcommand{\Initialization}{\item[\algorithmicinitialization]}   

\renewcommand{\eqref}[1]{Eq.~(\textup{\ref{#1}})}

\title{FedRE: A Representation Entanglement Framework for Model-Heterogeneous Federated Learning}

\author{
Yuan Yao$^{1}$ \hspace{0.5em}
Lixu Wang$^{2}$ \hspace{0.5em}
Jiaqi Wu$^{3,\dagger}$ \hspace{0.5em}
Jin Song$^{4}$ \hspace{0.5em}
Simin Chen$^{5}$ \hspace{0.5em}
Zehua Wang$^{6}$ \\
Zijian Tian$^{7}$ \hspace{0.5em}
Wei Chen$^{8}$ \hspace{0.5em}
Huixia Li$^{9}$ \hspace{0.5em}
Xiaoxiao Li$^{6}$
\\
$^{1}$Teleinfo, CAICT
$^{2}$Nanyang Technological University
$^{3}$Tsinghua University \\
$^{4}$Nanjing University of Posts and Telecommunications 
$^{5}$University of Texas at Dallas \\
$^{6}$University of British Columbia
$^{7}$China University of Mining and Technology-Beijing \\
$^{8}$China University of Mining and Technology
$^{9}$Beijing Jiaotong University
\\
$^{\dagger}$Corresponding author
}

\begin{document}

\maketitle

\begin{abstract}

Federated learning (FL) enables collaborative training across clients while preserving privacy. While most existing FL methods assume homogeneous model architectures, client heterogeneity in both data and resources makes this assumption impractical, thus motivating model-heterogeneous FL. To address this problem, we propose Federated Representation Entanglement (FedRE), a framework built upon a novel form of client knowledge termed entangled representation. Specifically, each client aggregates its local representations into a single entangled representation using normalized random weights, and then applies the same weights to integrate the corresponding one-hot label encodings into an entangled-label encoding. Both are subsequently uploaded to the server to train a global classifier. During training, each entangled representation is supervised across categories via its entangled-label encoding, while random weights are re-sampled at each round to introduce diversity, alleviating overconfidence in the global classifier and yielding smoother decision boundaries. Moreover, each client uploads a single entangled representation along with its entangled-label encoding, mitigating the risk of representation inversion attacks and reducing communication overhead. Extensive experiments demonstrate that FedRE achieves an effective trade-off among model performance, privacy protection, and communication overhead. The codes are available at 
\url{https://github.com/AIResearch-Group/FedRE}.

\end{abstract}

\section{Introduction}
\label{sec:intro}

\begin{figure}[t]
 \centering
 \includegraphics[width=\columnwidth]{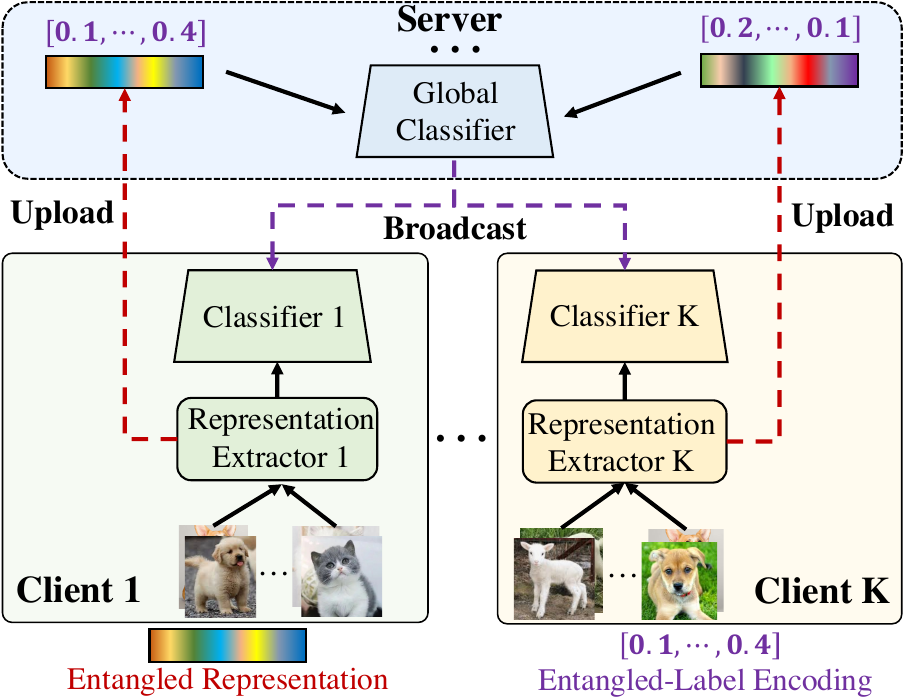}
 \caption{FedRE framework. Each client maintains a local model consisting of a representation extractor and a classifier. The client’s local representations and their corresponding one-hot label encodings are integrated into a single entangled representation and entangled-label encoding, respectively, which are then uploaded to the server for training the global classifier.}
 \label{fig:FedREworkflow}
 \vspace{-2ex}
\end{figure}

Federated learning (FL) \cite{mcmahan2017communication,yang2019federated} is a collaborative learning paradigm that aggregates client knowledge (\textit{e.g.}, \textit{model parameters}) from multiple clients while preserving privacy. Numerous FL methods have been developed and applied in various fields, such as healthcare \cite{antunes2022federated,zhang2025towards} and the Internet of Things \cite{nguyen2021federated,fan2024taking}. Most existing FL studies \cite{mcmahan2017communication,dong2022federated,li2023revisiting,pang2023collaborative,zhang2024improving,wang2025federated} assume that the architectures of local models across clients are homogeneous. In practice, however, assuming the same model architecture for all clients is unrealistic due to differences in sample distribution, hardware, and computational capabilities. 
Moreover, client model architectures may be private and not accessible to the server or other clients.
Those issues motivate the challenging setting of \textit{model-heterogeneous FL}~\cite{ye2023heterogeneous}, where representation extractors may vary across clients while classifiers remain homogeneous. Hence, directly aggregating all model parameters becomes infeasible.

To tackle this dilemma, existing model-heterogeneous FL studies have explored various forms of client knowledge, such as \textit{representations} \cite{makhija2022architecture}, \textit{logits} \cite{itahara2021distillation}, \textit{small-models} \cite{yi2024federated,wu2024fiarse}, \textit{classifiers} \cite{liang2020think}, or \textit{prototypes} (\textit{i.e.}, category means) \cite{tan2022fedproto,yi2023fedgh,huang2023rethinking,wang2024taming}, from clients.
While representations, logits, and small-models can effectively encode high-level client knowledge, uploading them to the server may introduce non-negligible communication overhead and potential privacy concerns, as such information could be exploited to reconstruct original samples by launching representation or model inversion attacks \cite{ulyanov2018deep,yin2020dreaming}.
As a lightweight alternative, uploading classifiers or prototypes reduces communication overhead and mitigates sample reconstruction risks. However, classifiers may inherit biases from local sample distributions, while prototypes capture category-level information with limited intra-class variability. This raises a question: ``\textit{For model-heterogeneous FL, is there a more effective, privacy-aware, and lightweight form of client knowledge?}''

To address this question, we propose a novel form of client knowledge, termed \textbf{\textit{entangled representation}}, which aggregates local representations across multiple categories into a single cross-category representation for each client. Building on this concept, we design a \textit{Federated Representation Entanglement} (FedRE) framework. 
As illustrated in \cref{fig:FedREworkflow}, each client aggregates its local representations into a single entangled representation using normalized random weights, and then applies the same weights to integrate the corresponding one-hot label encodings into an entangled-label encoding. Both are subsequently uploaded to the server to train the global classifier. During training, the entangled-label encodings provide cross-category supervision signals, and the per-round resampling of random weights introduces diversity, alleviating overconfidence in the global classifier and yielding smoother decision boundaries. 
Furthermore, entangled representations mitigate the risk of representation inversion attacks \cite{ulyanov2018deep} by blending cross-category information to obscure individual samples, while uploading only one representation per client reduces communication overhead.
As a result, entangled representations offer an effective, privacy-aware, and lightweight form of client knowledge.

The main contributions of this paper are threefold. 
\begin{itemize}
\item We introduce \textit{entangled representations}, a novel form of client knowledge.
\item We propose the FedRE framework, which supports flexible instantiations with different representation entanglement mechanisms.
\item Extensive experiments show that FedRE achieves an effective trade-off among model performance, privacy protection, and communication overhead. 
\end{itemize}

\section{Related Work}
\label{sec:relatedwork}

Existing FL methods can be broadly categorized into model-heterogeneous and model-homogeneous methods.

\textbf{Model-heterogeneous FL method} handles heterogeneity in both local models and sample distributions. Due to model heterogeneity across clients, direct parameter aggregation is not feasible. Thus, most studies have explored various forms of client knowledge.
For example, DS-FL \cite{itahara2021distillation} integrates the local logits from different clients into the global logit on the server. FedHeNN \cite{makhija2022architecture} aligns the representations extracted by distinct local models using a common representation alignment dataset. LG-FedAvg \cite{liang2020think} aggregates the classifiers from distinct clients on the server. Additionally, FedProto \cite{tan2022fedproto}, FPL \cite{huang2023rethinking}, FedGH \cite{yi2023fedgh}, and FedTGP \cite{zhang2024fedtgp} treat prototypes as a form of client knowledge. Specifically, FedProto \cite{tan2022fedproto} averages the local prototypes of each category across clients, and FPL employs a clustering strategy to derive unbiased global prototypes. FedTGP optimizes trainable global prototypes, while FedGH \cite{yi2023fedgh} utilizes prototypes to train the global classifier on the server. Moreover, several studies \cite{Zhu2021Data,Wu2022Communication} focus on knowledge distillation. For instance, FedGen \cite{Zhu2021Data} learns a global generator to augment the training samples for local models, while FedKD \cite{Wu2022Communication} distills a global student model to assist the learning of local models. Furthermore, another line of research \cite{wu2024fiarse,yi2024federated} uses homogeneous small-models. An example is FedMRL \cite{yi2024federated}, which facilitates inter-client knowledge aggregation via a shared small-model.

\textbf{Model-Homogeneous FL methods} handle heterogeneity in sample distributions while maintaining homogeneous local models. Most studies \cite{mcmahan2017communication,zhang2023fedala,ma2022layer,li2023revisiting,deng2022improving,deng2021fair, wang2021addressing} focus on aggregating all model parameters.
For example, FedAvg \cite{mcmahan2017communication} aggregates all parameters from distinct clients. Based on FedAvg, FedAvgDBE \cite{Zhang2023Eliminating}, FedFN \cite{kim2023fedfn}, and FedDecorr \cite{shi2023understanding} alleviates the representation bias issue in local models.
Another example is FedALA \cite{zhang2023fedala}, which adaptively integrates the global and local models to align with the local objective. Additionally, several studies \cite{sun2023fedperfix,collins2021exploiting,oh2021fedbabu,pillutla2022federated} aggregate partial model parameters. For instance, FedRep \cite{collins2021exploiting} aggregates the representation extractors of local models. Moreover, FedBABU \cite{oh2021fedbabu}, SphereFed \cite{dong2022spherefed}, FedETF \cite{li2023no}, and FedDr+ \cite{kim2025feddrplus} focus on improving representation learning and classifier design.
Furthermore, FedMP \cite{huang2025soft} alleviates conflicts between local and global representations via a multi-path framework, FedGPS \cite{yang2025fedgps} mitigates data heterogeneity via statistical rectification, and FedPhoenix \cite{wu2025rising} improves model generalization via dynamic parameter resetting.

\section{Methodology}
\label{sec:methodology}

In this section, we present the proposed FedRE framework. We begin by formalizing the problem addressed in this work. Consider $K$ clients and a server, where each client $k$ holds a private local dataset $\mathcal{D}_k = \{(\mathbf{x}_i^k, \mathbf{y}_i^k)\}_{i=1}^{n_k}$, with $\mathbf{x}_i^k$ denoting the $i$-th input sample and $\mathbf{y}_i^k$ denoting its one-hot label encoding over $C$ categories. Each client $k$ maintains a local model defined as $h_k(\bm{\theta}_k; \mathbf{x}) = f_k(\bm{\omega}_k; \bm{g}_k(\bm{\phi}_k; \mathbf{x}))$, where $\bm{g}_k$ is the representation extractor parameterized by $\bm{\phi}_k$, $f_k$ is the local classifier parameterized by $\bm{\omega}_k$, and $\bm{\theta}_k = \{\bm{\phi}_k, \bm{\omega}_k\}$. 
Representation extractors may vary across clients to accommodate architectural heterogeneity, while all local classifiers share the same architecture. The goal is to train client models on $\{\mathcal{D}_k\}_{k=1}^K$ to achieve high average accuracy across clients, while alleviating the risk of representation inversion attacks \cite{ulyanov2018deep} and reducing communication overhead. Next, we elaborate on FedRE’s motivation, workflow, and analyses.

\subsection{Motivation}

\begin{figure*}[t]
 \centering
 \includegraphics[width=\textwidth]{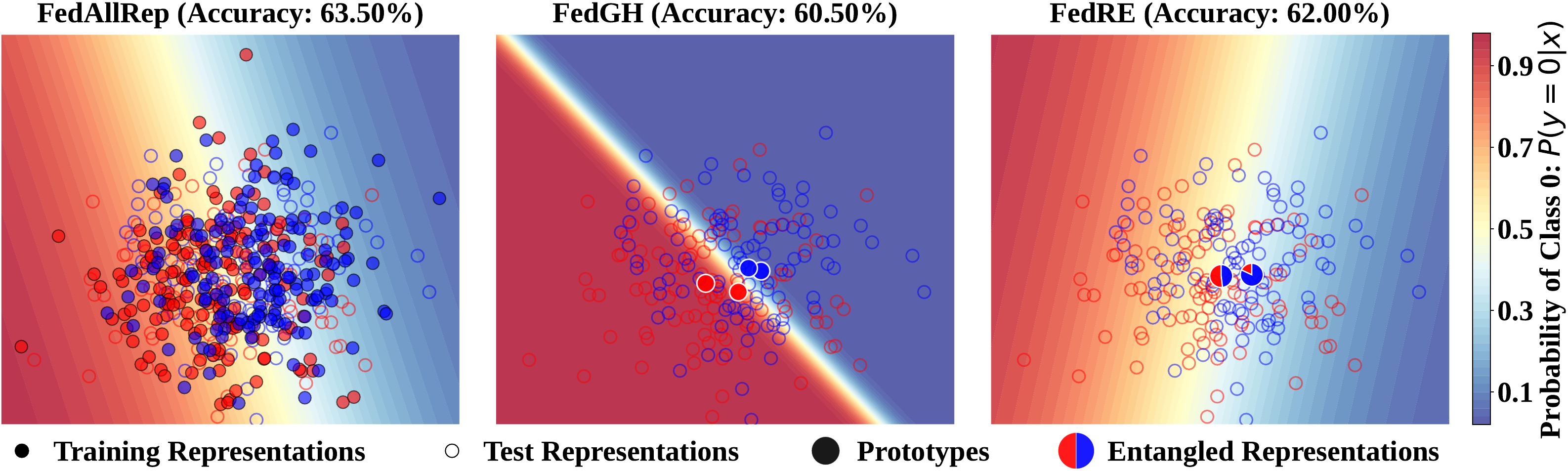}
 \caption{A toy experiment is conducted on 300 training and 200 test two-dimensional samples distributed across two clients. Red and blue denote predicted probabilities for classes 0 and 1, respectively, with darker colors indicating higher prediction confidence. FedAllRep uploads 300 representations from all clients to train the global classifier and achieves the best performance (63.50\%). FedGH uploads 4 prototypes, yielding sharper decision boundaries with abrupt color transitions and slightly lower performance (60.50\%). In contrast, FedRE uploads only 2 entangled representations that provide natural cross-category supervision, leading to smoother decision boundaries with gradual color transitions while maintaining competitive performance (62.00\%).}
\label{fig:Motivation}
\end{figure*}

As aforementioned, a key obstacle in model-heterogeneous FL lies in the heterogeneity of local representation extractors, which prevents direct parameter aggregation as done in FedAVG \cite{mcmahan2017communication}. To address this challenge, a promising direction is to incorporate client knowledge from multiple clients to train a high-quality global classifier while preserving privacy. 
Such a classifier leverages cross-client knowledge and improves local model performance upon deployment to clients.
A vanilla method, FedAllRep, uploads all sample representations to the server to train the global classifier. This ensures model performance by leveraging all representations (as exemplified in the left of \cref{fig:Motivation}), but it poses risks of leaking original samples via representation inversion attacks \cite{ulyanov2018deep} and introduces non-negligible communication overhead.
To alleviate this issue, FedGH \cite{yi2023fedgh} constructs client knowledge in the form of per-category prototypes to train the global classifier, thereby reducing communication cost and mitigating the risk of representation inversion attacks. Those prototypes capture category-representative information, and training on them may bias the global classifier toward prototypes, resulting in sharper decision boundaries (as exemplified in the middle of \cref{fig:Motivation}).

This limitation motivates the design of \textit{entangled representations}, a novel form of client knowledge that, unlike prototypes, integrates representations from distinct categories. 
Specifically, each client assigns a normalized random weight to each local representation. 
Those weights are then used to aggregate both the local representations and the corresponding one-hot label encodings into an entangled representation and an entangled-label encoding, respectively. Both are subsequently uploaded to the server to train the global classifier.
During training, the entangled-label encodings provide cross-category supervision signals, and the weights are re-sampled in each communication round to introduce diversity. This enables the global classifier avoid overconfidence in any single category and promotes smoother decision boundaries
(as exemplified in the right of \cref{fig:Motivation}).
Consequently, training the global classifier on entangled representations may lead to better performance than using prototypes. Furthermore, the entangled representation increases the difficulty of representation inversion attacks by blending all local representations into a single entangled representation, while uploading only one entangled representation per client further reduces communication overhead.
As a result, entangled representations provide effective, privacy-aware, and lightweight client knowledge, forming the foundation of FedRE. Next, we detail the FedRE.

\subsection{FedRE}

In FedRE, each client has a local model comprising a representation extractor and a classifier, as depicted in \cref{fig:FedREworkflow}. The FedRE workflow consists of three main steps: (i) local model update; (ii) representation entanglement and upload; and (iii) global classifier update and broadcast.

\subsubsection{Local Model Update}
Similar to vanilla FL methods such as FedAvg \cite{mcmahan2017communication}, FedRE requires each client to update its local model to effectively learn from local samples. To this end, the optimization objective for the $k$-th client is formulated by
\begin{equation}
\label{theta_k}
\min_{\bm{\theta}_k} \frac{1}{n_k} \sum_{(\mathbf{x}_i^k, \mathbf{y}_i^k) \in \mathcal{D}_k} \mathcal{L}_{ce} \big[ h_k (\bm{\theta}_k; \mathbf{x}_i^k), \mathbf{y}_i^k \big],
\end{equation}
where $\mathcal{L}_{ce} (\cdot, \cdot)$ denotes the cross-entropy loss.

\subsubsection{Representation Entanglement and Upload} 
\label{sec:RE}

We now describe the representation entanglement process. Each client first applies \textit{Representation Mapping} (RM) to its local representations, producing representations with consistent dimensionality for global classifier training. It then generates an entangled representation and an entangled-label encoding via \textit{Representation Entanglement} (RE):
\begin{equation}
\label{r_k_general}
\widetilde{\mathbf{r}}_k = \sum_{i=1}^{|\mathcal{D}_k|} w_i^k \text{RM} \big[ \bm{g}_k(\bm{\phi}_k; \mathbf{x}_i^k) \big], \widetilde{\mathbf{y}}_k = \sum_{i=1}^{|\mathcal{D}_k|} w_i^k \mathbf{y}_i^k,
\end{equation}
where $w_i^k \in [0, 1]$ denotes a normalized random weight assigned to sample $\mathbf{x}_i^k$. 
Then, each client uploads its entangled representation along with the corresponding entangled-label encoding, \textit{i.e.}, $\widetilde{\mathcal{R}} = \{ (\widetilde{\mathbf{r}}_k, \widetilde{\mathbf{y}}_k) \}_{k=1}^K$, to the server for training the global classifier.
Note that \textit{Eq.~(\ref{r_k_general}) defines a flexible framework supporting various RM and RE mechanisms}, as further analyzed in \textbf{Q6} and \textbf{Q7} of \Cref{sec:analysis}.

\subsubsection{Global Classifier Update and Broadcast} 
Upon receiving $\widetilde{\mathcal{R}} = \{ (\widetilde{\mathbf{r}}_k, \widetilde{\mathbf{y}}_k) \}_{k = 1}^K$, the server uses these pairs to train the global classifier. 
Accordingly, the server's optimization objective is formulated as
\begin{equation}
\label{loss_gloabl_classifier}
\min_{\bm{\omega}} \sum_{k = 1}^K \mathcal{L}_{ce} \big[ f(\bm{\omega}; \widetilde{\mathbf{r}}_k), \widetilde{\mathbf{y}}_k \big],
\end{equation}
where $f (\bm{\omega}; \cdot)$ denotes the global classifier parameterized by $\bm{\omega}$. 
Since entangled-label encodings provide cross-category supervision signals, minimizing Eq.~(\ref{loss_gloabl_classifier}) may encourage the global classifier to account for multiple categories and learn smoother decision boundaries.
Finally, the server broadcasts the updated global classifier to the clients for the next round.
With the above update process, the FedRE framework is summarized in \cref{FedRE}.

\begin{algorithm}
\caption{FedRE Framework}
\label{FedRE}
\begin{algorithmic}[1]
\Require $K$ clients with local datasets
\Ensure Trained local models of the clients
\Initialization 
Randomly initialize the global classifier and the local models
\For{$t = 0$ to $T - 1$}
    \Statex \textsc{// \textbf{Client Update:}}
    \For{each client $k$ in parallel}
        \State Receive the global classifier from the server to replace the local classifier
        \State Perform local fine-tuning on client $k$
        \State Apply RM to align local representations to a consistent dimensionality
        \State Apply RE to construct entangled representation-label pairs $(\widetilde{\mathbf{r}}_k, \widetilde{\mathbf{y}}_k)$ and upload them to the server
    \EndFor
    \Statex \textsc{// \textbf{Server Update:}}
    \State Update the global classifier using the uploaded representation-label pairs
    \State Broadcast the updated global classifier to the clients
\EndFor
\end{algorithmic}
\end{algorithm}

\subsection{Analysis}

\subsubsection{Computational Complexity of RE}

We analyze the computational complexity of RE on the client side. Let $\mathbf{R} \in \mathbb{R}^{n \times d}$ be the representation matrix, where $n$ is the number of samples and $d$ is the representation dimensionality, and let $\mathbf{Y} \in \mathbb{R}^{n \times C}$ denote the corresponding one-hot label matrix, where $C$ is the number of categories. Let $\mathbf{w} \in \mathbb{R}^{n}$ be the normalized weight vector. The entangled representation and entangled-label encoding are computed as $\widetilde{\mathbf{r}} = \mathbf{R}^\top \mathbf{w}$, and $\widetilde{\mathbf{y}} = \mathbf{Y}^\top \mathbf{w},$
respectively, yielding a total computational complexity of
$\mathcal{O}\big(n(d + C)\big).$

\subsubsection{Comparison with Related Studies}
\label{sec:comparison}

\textbf{Comparison with FL Methods}. FedRE differs from prototype-based methods (\textit{e.g.}, FedProto, FedGH) and representation-based methods (\textit{e.g.}, FedAllRep) in two key aspects: (1) FedRE uploads one entangled representation per client, reducing communication overhead and mitigating the risk of representation inversion attacks. (2) FedRE entangles cross-category representations into a unified one via random weighting, yielding better generalization than FedProto and FedGH and performance comparable to FedAllRep.

\noindent\textbf{Distinction from Mixup}. Unlike Mixup \cite{zhang2018mixup,verma2019manifold}, a data augmentation technique, FedRE is designed to address the challenges of FL and aims to balance model performance, privacy protection, and communication overhead. Specifically, while Mixup performs pairwise interpolation, FedRE aggregates all local representations within a client into a single entangled representation. Furthermore, the randomized weighting mechanism in FedRE generates diverse entangled representations across communication rounds, improving generalization under statistical heterogeneity across clients.

\subsubsection{Convergence Analysis of FedRE}

Empirically, as shown in \Cref{Fig:accu} and \Cref{appdix:Fig:accu} (Appendix~\ref{appendix:Model-HomogeneousSettingHeterogeneityAnalysis}), the performance of FedRE improves steadily and then stabilizes across communication rounds, indicating stable optimization in practice. From a theoretical perspective, the training process can be interpreted as an optimization procedure influenced by additional perturbations arising from local classifier replacement and local representation entanglement. A fully rigorous non-convex convergence analysis, in line with existing analyses (\textit{e.g.}, \cite{tan2022fedproto,yi2024federated}), is non-trivial due to these additional components, and we leave it for future work.

\section{Experiments}
\label{sec:Experiment}

In this section, we evaluate the proposed FedRE.

\subsection{Experimental Setup}
\label{sec:ExperimentalSetup}

\textbf{Datasets and Baselines}. We use three benchmark datasets: CIFAR-10 \cite{krizhevsky2009learning}, CIFAR-100 \cite{krizhevsky2009learning}, and TinyImageNet \cite{le2015tiny}. 
Moreover, we compare FedRE with eight state-of-the-art model-heterogeneous FL methods, including LG-FedAvg \cite{liang2020think}, FedGH \cite{yi2023fedgh}, FedKD \cite{Wu2022Communication}, FedGen \cite{Zhu2021Data}, FedProto \cite{tan2022fedproto}, FPL \cite{huang2023rethinking}, FedMRL \cite{yi2024federated}, and FedTGP \cite{zhang2024fedtgp}. We also include a Local method, where each client trains its model independently without communication.

\noindent\textbf{Model-Heterogeneous Settings}. We configure 10 clients with 10 heterogeneous architectures spanning diverse model families, including a four-layer CNN \cite{Zhang2023Eliminating}, MobileNetV2 \cite{sandler2018mobilenetv2}, GoogLeNet \cite{szegedy2015going}, five ResNet variants (ResNet-18, ResNet-34, ResNet-50, ResNet-101, ResNet-152) \cite{he2016deep}, and two Vision Transformer (ViT) models (ViT-B/16 and ViT-B/32) \cite{han2022survey}.

\noindent\textbf{Statistic-Heterogeneous Settings}. We follow \cite{Zhang2023Eliminating} and adopt both practical (PRA) \cite{lin2020ensemble,li2021model} and pathological (PAT) \cite{shamsian2021personalized} settings to simulate statistical heterogeneity among clients. In the PRA setting, samples are distributed across clients using a Dirichlet distribution \cite{lin2020ensemble} with a parameter $\alpha$, which is set to 0.1 by default across all datasets. In the PAT setting, each client is assigned samples from 2, 10, and 20 categories, drawn from a total of 10, 100, and 200 categories in CIFAR-10, CIFAR-100, and Tiny-ImageNet, respectively, with varying sample sizes. 

\noindent\textbf{Implementation Details}. We implement FedRE based on the PFLlib and HtFLlib frameworks \cite{zhang2025pfllib,zhang2025htfllib}, with 10 clients participating by default. 
The dataset is first partitioned across clients under statistical-heterogeneous settings, and each client's data is further split into a local training set (75\%) and a local test set (25\%).
We evaluate three RM operations and select \textit{Average Pooling} (AP) as the default (see \textbf{Q6} in \Cref{sec:analysis}), which averages representations across spatial regions into a unified dimensional vector for all clients. We further design five RE mechanisms and choose \textit{Random Average Prototype} (RAP) as the default (see \textbf{Q7} in \Cref{sec:analysis}).
In RAP, each client first computes prototypes and then aggregates both the prototypes and the corresponding one-hot label encodings using the same normalized random weights, yielding a single entangled representation and its corresponding entangled-label encoding.
Furthermore, we use SGD optimizers for both server and client updates, with task-dependent learning rates and batch sizes. The detailed experimental setup is provided in Appendix~\ref{appendix:DetailedExperimentalSetup}. All experiments are conducted on NVIDIA A800 GPUs.

\noindent\textbf{Evaluation Metrics.} We evaluate model performance by calculating the classification accuracy on the test set, averaged over all client models. To ensure a fair comparison, we report the average classification accuracy of the final round after 100 communication rounds, calculated across three random experiments, along with its standard deviation. To evaluate the robustness against representation inversion attacks, we use Peak Signal-to-Noise Ratio (PSNR) \cite{schlett2022face} and Mean Squared Error (MSE) between the reconstructed and original images. Lower PSNR and higher MSE indicate stronger privacy protection. Communication Overhead is measured by the total number of transmitted numerical scalars (\textit{i.e.}, parameters or representations) per round, where upload overhead represents the total scalars uploaded by all clients, and broadcast overhead represents the total scalars distributed from the server to all clients.

\subsection{Main Experiments}

\begin{figure}[t]
   \centering
   \subfloat[PRA]{
       \label{subfig:pra}
       \includegraphics[width=0.489\linewidth]{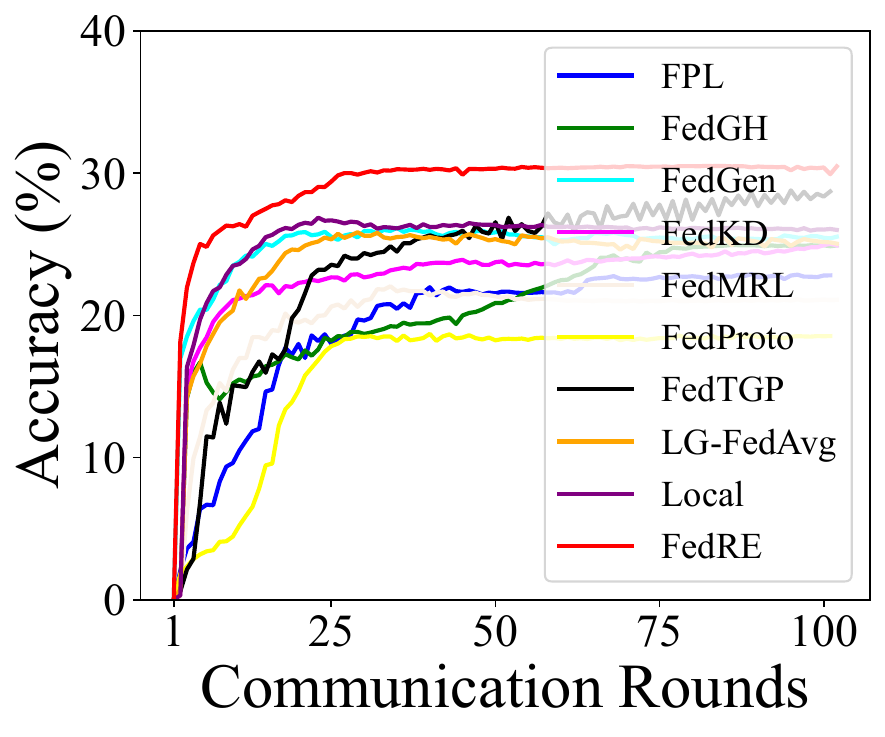}
   }
   \hspace{-2.7ex}
   \subfloat[PAT]{
       \label{subfig:pat}
       \includegraphics[width=0.489\linewidth]{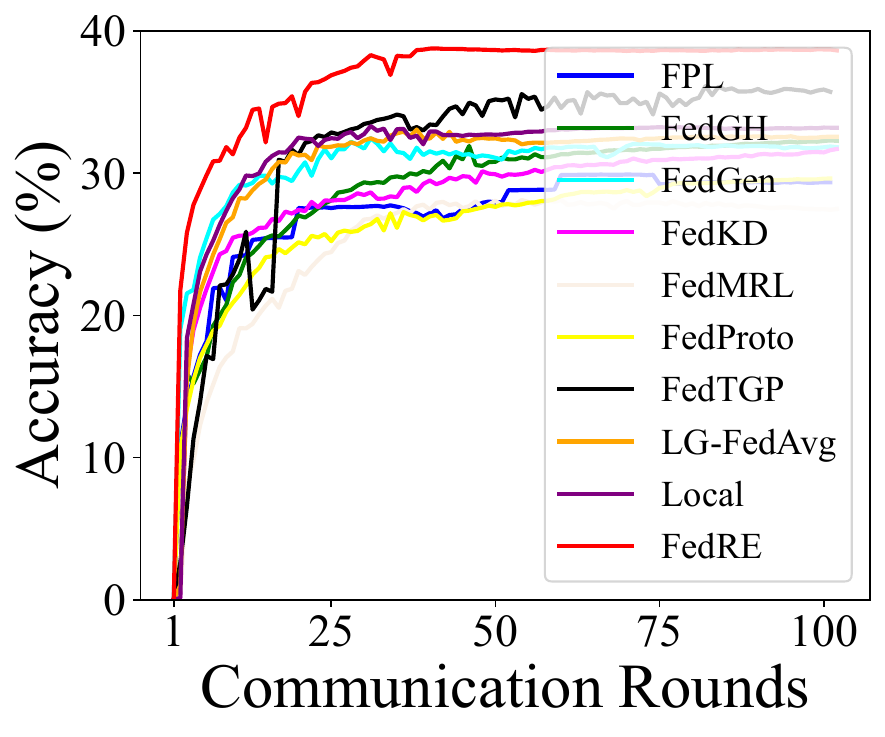}
   }
   \caption{Accuracy (\%) comparison between distinct communication rounds on the TinyImageNet dataset in the model-heterogeneous setting.}
   \label{Fig:accu}
   \vspace{-2ex}
\end{figure}

\textbf{Q1: How does FedRE perform in model-heterogeneous settings?} 
The results under the model-heterogeneous FL setting are listed in \Cref{tab:model-heterogeneous}. We have several insightful observations. (1) FedRE outperforms all the baselines across various scenarios. In particular, FedRE achieves an accuracy on TinyImageNet that surpasses those of LG-FedAvg, FedGH, and FedKD by 6.26\%, 6.54\%, and 6.79\%, respectively, under the PAT setting. Also, several methods do not exceed the performance of Local, indicating that the scenarios are challenging. 
(2) FedRE performs better than FedGH, suggesting that entangled representations may be more effective than prototypes for training the global classifier.
(3) LG-FedAvg is inferior to FedRE, demonstrating that optimizing the global classifier with entangled representations is superior to directly aggregating local classifiers.
Also, Figures~\ref{Fig:accu}(a)-(b) show that FedRE achieves rapid performance gains on TinyImageNet in the early rounds and then stabilizes, indicating stable convergence behavior.
Moreover, we evaluate FedRE under various statistical-heterogeneous settings in \textbf{Q4} of \Cref{sec:analysis}, including large-scale participation (\textit{i.e.}, \textbf{100 clients}).

\noindent\textbf{Q2: Can entangled representations effectively mitigate the risk of representation inversion attacks?}
We launch the representation inversion attacks \cite{ulyanov2018deep} to reconstruct original samples from representations, prototypes, and entangled representations, respectively. Detailed attack settings are provided in Appendix~\ref{appendix:DetailedExperimentalSetup}. \cref{fig:privacy} illustrates the reconstruction results for sample images from TinyImageNet. We can make several insightful observations. (1) Most image contours are reconstructed from the representations, indicating their vulnerability to representation inversion attacks. (2) Some category information, such as the presence of a \textit{fish}, is leaked through reconstructed prototypes, as prototypes encapsulate representative category information. (3) The reconstructed images from entangled representations reveal no identifiable information. This is because entangled representations combine information across different categories, making it difficult to reconstruct individual samples. In addition, the PSNR values for images reconstructed from representations, prototypes, and entangled representations are 12.89, 10.25, and 9.66, with corresponding MSE values of 4514.91, 6992.04, and 7781.87. Those results suggest that entangled representations tend to produce lower PSNR and higher MSE, which mitigates the risk of representation inversion attacks.

\begin{table*}[t]
\centering
\caption{Accuracy (\%) comparison on three datasets under the model-heterogeneous setting. In each column, the best results are \textbf{bolded}, and the second-best results are \underline{underlined}.}
\setlength{\tabcolsep}{1ex}
\resizebox{\textwidth}{!}{%
\begin{tabular}{cccccccc}
\toprule
\multirow{2}{*}{\textbf{Method}} & \multicolumn{3}{c}{\textbf{PRA}} & \multicolumn{3}{c}{\textbf{PAT}} & \multirow{2}{*}{\textbf{Average}} \\
\cmidrule(lr){2-4} \cmidrule(lr){5-7}
& \textbf{CIFAR-10} & \textbf{CIFAR-100} & \textbf{TinyImageNet} & \textbf{CIFAR-10} & \textbf{CIFAR-100} & \textbf{TinyImageNet} & \\
\midrule
LG-FedAvg \cite{liang2020think} & 80.90 $\pm$ 0.17 & \underline{41.96 $\pm$ 0.03} & 25.16 $\pm$ 0.42 & 85.35 $\pm$ 0.25 & \underline{58.24 $\pm$ 0.33} & 32.26 $\pm$ 0.28 & 53.98 \\
FedGH \cite{yi2023fedgh} & 78.66 $\pm$ 0.34 & 40.91 $\pm$ 0.26 & 25.04 $\pm$ 0.11 & \underline{85.43 $\pm$ 0.03} & 58.07 $\pm$ 0.33 & 31.98 $\pm$ 0.29 & 53.35 \\
FedKD \cite{Wu2022Communication} & 80.79 $\pm$ 0.38 & 41.33 $\pm$ 0.25 & 25.39 $\pm$ 0.36 & 84.03 $\pm$ 0.17 & 55.61 $\pm$ 0.10 & 31.73 $\pm$ 0.20 & 53.15 \\
FedGen \cite{Zhu2021Data} & 81.16 $\pm$ 0.12 & 41.46 $\pm$ 0.10 & 25.45 $\pm$ 0.19 & 84.88 $\pm$ 0.18 & 57.87 $\pm$ 0.67 & 31.96 $\pm$ 0.21 & 53.80 \\
FedProto \cite{tan2022fedproto} & 78.36 $\pm$ 0.52 & 35.00 $\pm$ 0.34 & 18.16 $\pm$ 0.08 & 83.81 $\pm$ 0.18 & 56.72 $\pm$ 0.11 & 29.61 $\pm$ 0.02 & 50.28 \\
FPL \cite{huang2023rethinking} & 77.40 $\pm$ 0.23 & 36.66 $\pm$ 0.45 & 22.64 $\pm$ 0.46 & 83.89 $\pm$ 0.38 & 53.21 $\pm$ 0.25 & 29.16 $\pm$ 0.19 & 50.49 \\
FedMRL \cite{yi2024federated} & 81.28 $\pm$ 0.05 & 34.41 $\pm$ 0.04 & 20.92 $\pm$ 0.09 & 83.30 $\pm$ 0.41 & 54.25 $\pm$ 0.21 & 27.37 $\pm$ 0.10 & 50.26 \\
FedTGP \cite{zhang2024fedtgp} & \underline{81.32 $\pm$ 0.47} & 35.89 $\pm$ 0.22 & \underline{28.70 $\pm$ 0.49} & 84.68 $\pm$ 0.27 & 54.67 $\pm$ 1.34 & \underline{35.64 $\pm$ 0.37} & 53.48 \\
Local & 81.20 $\pm$ 0.05 & 41.57 $\pm$ 0.10 & {25.81 $\pm$ 0.15} & 84.68 $\pm$ 0.07 & 57.96 $\pm$ 0.12 & {33.02 $\pm$ 0.14} & \underline{54.04} \\
FedRE & \textbf{82.60 $\pm$ 0.01} & \textbf{46.36 $\pm$ 0.09} & \textbf{30.48 $\pm$ 0.13} & \textbf{86.20 $\pm$ 0.14} & \textbf{62.56 $\pm$ 0.32} & \textbf{38.52 $\pm$ 0.08} & \textbf{57.79} \\
\bottomrule
\end{tabular}%
}
\label{tab:model-heterogeneous}
\vspace{-3ex}
\end{table*}

\noindent\textbf{Q3: What is the communication overhead of FedRE?} We conduct communication overhead experiments on the CIFAR-100 dataset under the PRA setting. As shown in \Cref{tab:CO}, FedRE achieves the lowest communication overhead during the upload phase, as it uploads only a single entangled representation along with its entangled-label encoding from each client to the server. During the broadcast phase, its overhead is comparable to that of classifier-based methods (\textit{e.g.}, LG-FedAvg) and prototype-based methods (\textit{e.g.}, FedProto). Those results imply that FedRE is effective in reducing communication overhead. More results are offered in Appendix~\ref{appendix:CommunicationOverheadEvaluation}.

\begin{figure}[t] 
    \centering 
    \subfloat[\footnotesize Original Images]{ 
         \includegraphics[width=1.35cm,height=1.35cm]{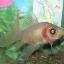}%
         \includegraphics[width=1.35cm,height=1.35cm]{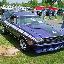}%
         \includegraphics[width=1.35cm,height=1.35cm]{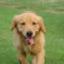}%
    }%
    \subfloat[\footnotesize Representations]{ 
         \includegraphics[width=1.35cm,height=1.35cm]{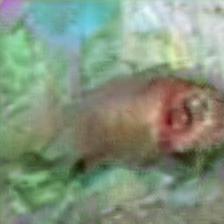}%
         \includegraphics[width=1.35cm,height=1.35cm]{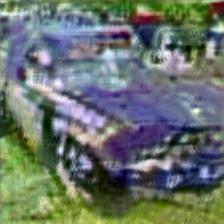}%
         \includegraphics[width=1.35cm,height=1.35cm]{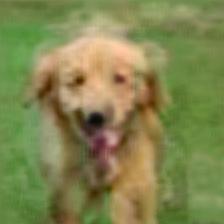}%
    }
    \\
    \subfloat[\footnotesize Prototypes]{ 
         \includegraphics[width=1.35cm,height=1.35cm]{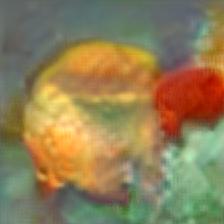}%
         \includegraphics[width=1.35cm,height=1.35cm]{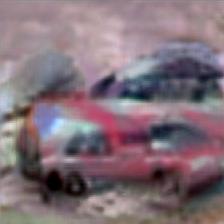}%
         \includegraphics[width=1.35cm,height=1.35cm]{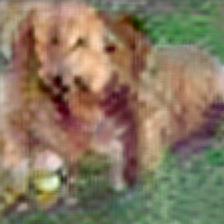}%
    }
    \subfloat[\footnotesize Entangled Representations]{ 
         \includegraphics[width=1.35cm,height=1.35cm]{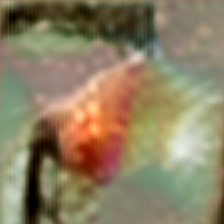}%
         \includegraphics[width=1.35cm,height=1.35cm]{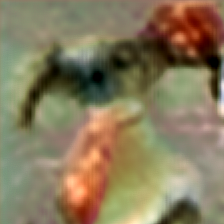}%
         \includegraphics[width=1.35cm,height=1.35cm]{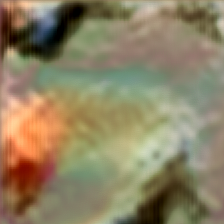}%
    }
    \caption{Comparison of privacy protection in restructuring results from representations, prototypes, and entangled representations on the TinyImageNet dataset.}
    \label{fig:privacy} 
\vspace{-2ex}
\end{figure}

\begin{table*}[t!] 
\centering 
\setlength{\tabcolsep}{0.3cm} 
\caption{{Communication overhead (\# Scalars $\times 10^3$) comparison on the CIFAR-100 dataset. In each row, the best results are \textbf{bolded}, and the second-best results are \underline{underlined}.}}
\label{tab:CO} 
\resizebox{\textwidth}{!}{%
\begin{tabular}{@{}c*{9}{c}@{}} 
\toprule
\textbf{Metric} & \textbf{LG-FedAvg} & \textbf{FedGH} & \textbf{FedKD} & \textbf{FedGen} & \textbf{FedProto} & \textbf{FedMRL} & \textbf{FedTGP} & \textbf{FPL} & \textbf{FedRE} \\
\midrule
Upload    & 513.00 & \underline{257.02} & 4234.28 & 9247.08 & \underline{257.02} & 8863.08 & \underline{257.02} & \underline{257.02} & \textbf{5.12} \\
Broadcast & \underline{513.00} & \textbf{512.00} & 4234.28 & \underline{513.00} & \textbf{512.00} & 8863.08 & \textbf{512.00} & 916.48 & \underline{513.00} \\
\bottomrule
\end{tabular}%
}
\end{table*}

\subsection{Analysis}
\label{sec:analysis}

\noindent\textbf{Q4: How does FedRE perform under different participation ratios with varying statistical heterogeneity?} We conduct experiments on the CIFAR-10 dataset under the PRA setting with \textit{partial} client participation and varying levels of statistical heterogeneity. Specifically, we adopt \textbf{100 clients} and set the participation rates to 10/100 and 20/100, while adjusting the Dirichlet distribution parameter $\alpha$ to 0.07 and 0.1, respectively, in the PRA setting. Furthermore, we follow \cite{lee2019synthesizing} and simulate the long-tail settings by modifying Imbalance Factors (IF) to 100 and 50, then set $\alpha$ to 0.07. 
We present the results in \Cref{table:participationIF} and highlight several observations.
(1) FedProto performs relatively poorly, potentially because limited same-category overlap across clients in highly heterogeneous settings weakens prototype aggregation.
(2) FedRE achieves the best performance in most scenarios, demonstrating its effectiveness under partial participation with highly heterogeneous distributions.
More results are provided in Appendix~\ref{appendix:statisticalheterogeneityanalysis}.

\begin{table*}[t]
\centering
\caption{Accuracy (\%) comparison for partial participation scenarios with varying statistical heterogeneity in the PRA setting on the CIFAR-10 dataset. Here, $\alpha$ is a Dirichlet distribution parameter, and IF denotes imbalance factors of the long-tail setting. In each column, the best results are \textbf{bolded}, and the second-best results are \underline{underlined}.}
\label{table:participationIF}
\setlength{\tabcolsep}{1.5pt}
\resizebox{\textwidth}{!}{
\begin{tabular}{c *{8}{c}}
\toprule
\multirow{2}{*}{\textbf{Method}}
&
\multicolumn{4}{c}{\textbf{Participation Rate = 10/100}} &
\multicolumn{4}{c}{\textbf{Participation Rate = 20/100}} \\
\cmidrule(lr){2-5} \cmidrule(lr){6-9}
& $\bm{\alpha=0.07}$ & $\bm{\alpha=0.1}$ & \textbf{IF=100,} $\bm{\alpha=0.07}$ & \textbf{IF=50,} $\bm{\alpha=0.07}$ &
$\bm{\alpha=0.07}$ & $\bm{\alpha=0.1}$ & \textbf{IF=100,} $\bm{\alpha=0.07}$ & \textbf{IF=50,} $\bm{\alpha=0.07}$ \\
\midrule
FedProto \cite{tan2022fedproto}
& 54.00 & 51.18 & 45.21 & 43.92 
& 56.90 & 55.47 & 47.08 & 44.68 \\
FedGH  \cite{yi2023fedgh}
& \underline{78.23} & \underline{76.87} & \textbf{67.30} & \underline{63.73} 
& \underline{80.57} & \underline{77.84} & \underline{65.56} & \underline{64.90} \\
FedRE  
& \textbf{81.17} & \textbf{79.56} & \underline{67.12} & \textbf{66.37} 
& \textbf{82.80} & \textbf{81.99} & \textbf{69.33} & \textbf{68.81} \\
\bottomrule
\end{tabular}
}
\vspace{-2ex}
\end{table*}

\noindent\textbf{Q5: What are the advantages of uploading a single entangled representation per client compared to uploading all representations?} To investigate the benefits of entangled representations, we compare FedRE with FedAllRep, which uploads all local representations to train the global classifier. \Cref{tab:FusedPrototype} reports the results on TinyImageNet under both PRA and PAT settings. As can be seen, FedRE achieves performance comparable to FedAllRep, suggesting that uploading a single entangled representation per client can effectively support the training of the global classifier. Moreover, FedRE effectively reduces communication overhead compared to uploading all representations.

\begin{table}[htbp]
   \centering
   \caption{Accuracy (\%) and communication overhead comparison on the TinyImageNet dataset. In each column, the best results are \textbf{bolded}, and the second-best results are \underline{underlined}.}
   \label{tab:FusedPrototype}
   \resizebox{\columnwidth}{!}{%
   \begin{tabular}{@{}c@{\hskip 2pt}c@{\hskip 2pt}c@{\hskip 2pt}c@{\hskip 2pt}c@{}}
   \toprule
   \multirow{2}{*}{\textbf{Method}} & \multicolumn{2}{c}{\textbf{PRA}} & \multicolumn{2}{c}{\textbf{PAT}} \\
   \cmidrule(lr){2-3} \cmidrule(lr){4-5}
   & \textbf{Accuracy} & \textbf{\# Scalars ($\times 10^3$)} & \textbf{Accuracy} & \textbf{\# Scalars ($\times 10^3$)} \\
   \midrule
   FedAllRep & \textbf{31.20} & \underline{42160.39} & \textbf{38.62} & \underline{42258.88} \\
   FedRE     & \underline{30.48} & \textbf{4118.48} & \underline{38.52} & \textbf{4118.48} \\
   \bottomrule
   \end{tabular}%
   }
\end{table}

\noindent\textbf{Q6: How effective are different RM operations?}
We evaluate three distinct RM operations: (1) Average Pooling (AP) averages representation values across spatial regions. (2) Max Pooling (MP) selects the maximum representation values across regions. (3) Fully Connected layer (FC) performs a learned aggregation operation on representations. \Cref{tab:RepresentationMapping} presents results on the CIFAR-100 dataset under both PRA and PAT settings. We observe that AP achieves favorable performance and is thus set as the default choice in FedRE.

\begin{table}[H]
   \centering
   \setlength{\tabcolsep}{0.6cm}
   \caption{Accuracy (\%) comparison between distinct RM operations on the CIFAR-100 dataset in both PRA and PAT settings. In each row, the best results are \textbf{bolded}, and the second-best results are \underline{underlined}.}
   \label{tab:RepresentationMapping}
   \resizebox{0.8\linewidth}{!}{%
   \begin{tabular}{@{}c*{3}{c}@{}}
       \toprule
       \textbf{Setting} & \textbf{AP} & \textbf{MP} & \textbf{FC}  \\
       \midrule
       PRA &  \textbf{46.36} & \underline{45.97} & 44.53  \\
       PAT &  \textbf{62.56} & \underline{61.93} & 60.19  \\
       \bottomrule
   \end{tabular}%
   }
   \vspace{-1ex}
\end{table}

\noindent\textbf{Q7: How effective are distinct RE mechanisms?}
We design five distinct RE mechanisms, with their mathematical formulations provided in Appendix~\ref{Appendix:RepresentationEntanglementDetails}. Below, we only outline how entangled representations are obtained, as entangled-label encodings are derived analogously from one-hot label encodings:
(1) Random Select Representation (RSR) randomly selects one representation from each client. 
(2) Vanilla Average Representation (VAR) averages all representations per client into a single representation, with equal weight assigned to each.
(3) Random Average Representation (RAR) entangles all representations per client into a single representation using a normalized weight vector, with elements randomly drawn from a Uniform distribution $\mathcal{U}(0, 1)$ and normalized to sum to one.
(4) Random Select Prototype (RSP) first calculates prototypes for each client and then randomly selects one prototype per client.
(5) Vanilla Average Prototype (VAP) calculates prototypes for each client and averages them into a single representation, where each prototype contributes equally.
(6) Random Average Prototype (RAP) calculates prototypes for each client and aggregates them into a single representation using a normalized weight vector, where each weight is randomly drawn from a Uniform distribution $\mathcal{U}(0, 1)$ and normalized to sum to one. \Cref{tab:EntangledMechanism} lists the results on the CIFAR-10 and CIFAR-100 datasets in the PRA setting. We have the following observations. 
(1) RSR performs the worst, as each client uploads only a randomly selected representation, which is insufficient to train the global classifier.
(2) RSP outperforms RSR, as the prototype aggregates all representations within a category, it is more representative than a single vanilla representation. 
(3) VAP and RAP outperform VAR and RAR, respectively, indicating that prototype-based entanglement yields better model performance. 
(4) RAP surpasses VAP, demonstrating that random weights for entanglement are more effective than equal weights.
Thus, we empirically choose RAP in the implementation of FedRE.

\begin{table}[htbp]
   \centering
   \caption{Accuracy (\%) comparison between distinct RE mechanisms on the CIFAR-10 and CIFAR-100 datasets in the PRA setting. In each row, the best results are \textbf{bolded}, and the second-best results are \underline{underlined}.}
   \resizebox{\columnwidth}{!}{%
   \begin{tabular}{@{}c*{6}{c}@{}}
       \toprule
       \textbf{Dataset} & \textbf{RSR} & \textbf{VAR} & \textbf{RAR} & \textbf{RSP} & \textbf{VAP} & \textbf{RAP} \\
       \midrule
       CIFAR-10 &  79.10 & 81.32 & 80.20 & 80.45 & \underline{81.42} & \textbf{82.60} \\
       CIFAR-100 & 40.41 & 44.88 & 43.19 & 43.25 & \underline{46.12} & \textbf{46.36} \\
       \bottomrule
   \end{tabular}%
   }
   \label{tab:EntangledMechanism}
\end{table}

\noindent\textbf{Q8: How do different distributions in RAP affect FedRE performance?}
As mentioned above, in RAP, we sample weights from a Uniform distribution $\mathcal{U}(0,1)$.
To examine the effect of the distribution choice, we replace it with a Laplace distribution $\mathcal{L}(0,1)$ and a Gaussian distribution $\mathcal{G}(0,1)$. The results on the CIFAR-10 and CIFAR-100 datasets under the PRA setting are reported in \Cref{tab:distribution}. We can observe that RAP achieves comparable performance across different distributions, with a minor advantage for the Uniform distribution. This indicates that FedRE is flexible in supporting diverse distribution configurations in RAP.

\begin{table}[htbp]
    \centering
    \caption{Accuracy (\%) comparison of different distributions in RAP on the CIFAR-10 and CIFAR-100 datasets in the PRA setting. In each row, the best results are \textbf{bolded}, and the second-best results are \underline{underlined}.}
    \begin{tabular}{@{}c*{3}{c}@{}}
        \toprule
        \textbf{Dataset} & \textbf{Gaussian} & \textbf{Laplace} & \textbf{Uniform} \\
        \midrule
        CIFAR-10 & \underline{82.39} & 82.32 & \textbf{82.60} \\
        CIFAR-100 & 45.51 & \underline{45.84} & \textbf{46.36} \\
        \bottomrule
    \end{tabular}%
    \label{tab:distribution}
\end{table}

\noindent\textbf{Q9: How effective is per-round random weight re-sampling in FedRE?}
In each communication round, each client performs random weight re-sampling (RS) for its local representations. To evaluate its effectiveness, we compare RS with a fixed-sampling (FS) variant, in which the weights are sampled only once at initialization and then reused in all subsequent rounds.
Experiments are conducted on a synthetic dataset consisting of 300 training and 200 test two-dimensional samples distributed across two simulated clients (same as the dataset used in \cref{fig:Motivation}), as well as on the CIFAR-100 dataset in the PRA setting with 10 clients.
As shown in \Cref{tab:FWvsDW}, RS achieves higher accuracy than FS across both datasets, which suggests the effectiveness of the per-round random weight re-sampling.

\begin{table}[htbp]
  \centering
  \caption{Accuracy (\%) comparison between re-sampling (RS) and fixed-sampling (FS) in FedRE on a synthetic dataset and the CIFAR-100 dataset. In each row, the best and second-best results are highlighted in bold and \underline{underline}, respectively.}
    \begin{tabular}{ccc}
    \toprule
    \textbf{Dataset} & \textbf{FS} & \textbf{RS}  \\
    \midrule
    Synthetic Dataset & \underline{41.50}  & \textbf{62.00}  \\
    CIFAR-100 & \underline{45.84}  & \textbf{46.36}  \\
    \bottomrule
    \end{tabular}%
  \label{tab:FWvsDW}%
\end{table}%

\noindent\textbf{Q10: Is the training cost introduced by RE during local training significant in FedRE?}
In FedRE, we introduce an RE mechanism during local training (LT). To evaluate its training cost, we compare two settings: LT without RE and LT with RE, where the former ablates the RE component. \Cref{tab:efficiency} reports the average training cost (in seconds) per round on CIFAR-10 and CIFAR-100, averaged across 10 clients and 6 communication rounds. As shown, LT with RE incurs only a slight additional cost compared to LT without RE, indicating that the RE mechanism introduces minor computational cost. This is mainly because RE separately aggregates representations and label encodings without additional gradient computations.

\begin{table}[htbp]
   \centering
   \caption{Comparison of the average time cost (in seconds) per round for local training (LT) in FedRE without and with RE. In each row, the best results are \textbf{bolded}, and the second-best results are \underline{underlined}.}
   \begin{tabular}{ccc}
   \toprule
   \textbf{Dataset} & \textbf{LT without RE} & \textbf{LT with RE} \\
   \midrule
   CIFAR-10 & \textbf{5.69} & \underline{5.78} \\
   CIFAR-100 & \textbf{5.70} & \underline{5.79} \\
   \bottomrule
   \end{tabular}%
   \label{tab:efficiency}%
\end{table}

\textbf{Additional Analysis}. Appendices~\ref{appendix:Model-HomogeneousSettingHeterogeneityAnalysis}-\ref{appendix:SmallClients} further verify the effectiveness of FedRE under both the model-homogeneous setting and a small number of clients.

\section{Conclusion}
\label{sec:Conclusion}

In this paper, we introduce the entangled representation as an effective, privacy-aware, and lightweight form of client knowledge. Building on this concept, we propose the FedRE framework for model-heterogeneous FL. In FedRE, each client first produces a single cross-category entangled representation along with its associated entangled-label encoding, which are then uploaded to the server for global classifier training. Experimental results demonstrate that FedRE achieves a well-balanced trade-off among model performance, privacy protection, and communication overhead. A promising direction for future work is to explore the applicability of entangled representations to a broader range of machine learning tasks.

{
\small
\bibliographystyle{ieeenat_fullname}
\bibliography{FedRE}
}

\clearpage
\onecolumn
\appendix

Additional details and results are provided in the appendices, including:

\begin{itemize}

\item Appendix~\ref{Appendix:RepresentationEntanglementDetails}: Mathematical details of various representation entanglement mechanisms.

\item Appendix~\ref{appendix:DetailedExperimentalSetup}: Detailed experimental setup.

\item Appendix~\ref{appendix:MoreExp}: Supplementary experimental results.

\end{itemize}

\section{Mathematical Details of Various Representation Entanglement Mechanisms}
\label{Appendix:RepresentationEntanglementDetails}

We now introduce the mathematical details of different RE mechanisms. The general form of RE, calculated from a single client's representation set $\mathcal{R} = \{ (\mathbf{r}_i, \mathbf{y}_i) \}_{i = 1}^n$, is formulated by
\begin{equation}
\label{eq:RE}
\widetilde{\mathbf{r}} = \sum_{i = 1}^n w_i \mathbf{r}_i,
\widetilde{\mathbf{y}} = \sum_{i = 1}^n w_i \mathbf{y}_i,
\end{equation}
where $w_i \in [0, 1]$ is the weight of $\mathbf{r}_i$, which is determined by different RE mechanisms as follows: 

\begin{itemize}[left=0pt]

\item \textbf{Random Select Representation} (RSR) randomly selects one representation from each client per global communication round. Thus, 
$w_i$ is formulated as
\begin{equation}
w_i = \begin{cases} 
1, & \text{if  } \mathbf{r}_i \text{  is selected} \\
0, & \text{otherwise}.
\end{cases}
\end{equation}

\item \textbf{Vanilla Average Representation} (VAR) averages all representations per client into a single representation, with equal weight assigned to each. Hence, $w_i$ is defined by
\begin{equation}
w_i = \frac{1}{n}, \forall i \in \{ 1, 2, \cdots, n \}.
\end{equation}

\item \textbf{Random Average Representation} (RAR) {entangles} representations per client into a single representation using a normalized weight vector, with elements randomly drawn from a Uniform distribution $\mathcal{U}(0, 1)$ and normalized to sum to one. Accordingly, $w_i$ is formulated as follows:
\begin{equation}
w_i = \frac{u_i}{\sum_{j = 1}^n u_j}, \text{where  } u_i \sim \mathcal{U} (0, 1).
\end{equation}

\item \textbf{Random Select Prototype} (RSP) first calculates prototypes for each client and then randomly selects one prototype per client in each global communication round. Therefore, $w_i$ is defined as
\begin{equation}
w_i = \begin{cases} 
\frac{1}{n_c}, & \text{if both the selected prototype and } \mathbf{r}_i \text{belong to category } c \\
0, & \text{otherwise},
\end{cases}
\end{equation}
where $n_c$ denotes the total number of samples belonging to category $c$.

\item \textbf{Vanilla Average Prototype} (VAP) calculates prototypes for each client and averages them into a single representation, where each prototype contributes equally. Thus, $w_i$ is calculated by
\begin{equation}
w_i = \frac{1}{C n_c}, \text{if  } \mathbf{r}_i \text{belongs to category } c,
\end{equation}
where $C$ is the total number of categories in the client.

\item \textbf{Random Average Prototype} (RAP) calculates prototypes for each client and aggregates them into a single representation using a normalized random weight vector, where each weight $u_c \sim \mathcal{U}(0,1)$, and the weights are normalized to sum to one. Using those weights, $w_i$ is defined as follows:

\begin{equation}
w_i = \frac{u_c}{n_c \sum_{j=1}^C u_j}, \quad \text{if } \mathbf{r}_i \text{ belongs to category } c.
\end{equation}

\end{itemize}

\section{Detailed Experimental Setup}
\label{appendix:DetailedExperimentalSetup}

\Cref{tab:experimentalsetup} provides a detailed description of the experimental setup used in this paper, covering statistical-heterogeneous settings, model training details, and model configurations. In addition, we detail the representation inversion attack setup. We assume a \textit{semi-honest} server with full access to the representation extractor $\bm{g}(\bm{\phi}; \cdot)$, attempting to reconstruct clients’ original samples via a representation inversion attack. Following~\cite{ulyanov2018deep}, given a target vector $\mathbf{r}$ (\textit{e.g.}, representation, prototype, or entangled representation), the server optimizes an inversion network $\bm{I}(\psi; \cdot)$ with fixed noise $\mathbf{z}$ by solving $\min_{\psi} \| \bm{g}(\bm{\phi}; \bm{I}(\psi; \mathbf{z})) - \mathbf{r} \|_2^2$. To ensure a fair comparison, we use a single four-layer CNN as the representation extractor and adopt the same inversion network as in~\cite{ulyanov2018deep} for all reconstruction experiments.

\begin{table}[htbp]
\centering
\caption{\itshape Detailed experimental setup used in this paper.}
\resizebox{\textwidth}{!}{%
\begin{tabular}{cc}
\toprule
\textbf{Statistic-heterogeneous Settings} & Practical setting (PRA); Pathological setting (PAT) \\
\midrule
\multirow{8}{*}{\textbf{Model Training Details}} & Local batch size: 64 (TinyImageNet), 32 (CIFAR-10 \& CIFAR-100) \\
& Local optimizer: SGD \\
& Local learning rate $\eta_l$: \\
& \quad 0.06 (Model-Heterogeneity with PRA \& PAT, CIFAR-10/100/TinyImageNet) \\
& \quad 0.01 (Model-Homogeneity with PRA \& PAT, CIFAR-100/TinyImageNet) \\
& \quad 0.007 (Model-Homogeneity with PRA, CIFAR-10) \\
& \quad 0.008 (Model-Homogeneity with PAT, CIFAR-10) \\
& Server batch size: 10; Server optimizer: SGD; Server learning rate: 0.01 \\
\midrule
\multirow{3}{*}{\textbf{Model Configurations}} & Local model in Model-Heterogeneity: CNN; MobileNetV2; GoogleNet; \\
& ResNet-18/34/50/101/152; ViT-B/16; ViT-B/32 \\
& Local model in Model-Homogeneity: CNN (CIFAR-10/100); ResNet-18 (TinyImageNet) \\
\bottomrule
\end{tabular}%
}
\vspace{-2ex}
\label{tab:experimentalsetup}
\end{table}

\section{Supplementary Experimental Results}
\label{appendix:MoreExp}

\subsection{Model-homogeneous FL Evaluation} 
\label{appendix:Model-HomogeneousSettingHeterogeneityAnalysis}

\begin{figure}[H]
    \centering
    \subfloat[PRA]{
    \includegraphics[width=0.35\columnwidth]{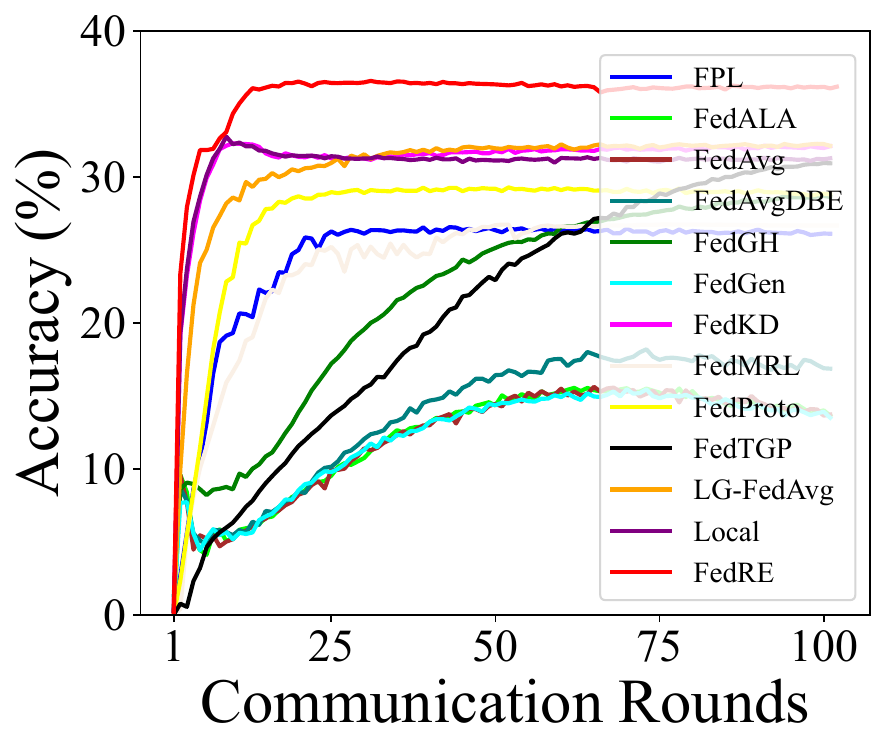}}
    \subfloat[PAT]{
    \includegraphics[width=0.35\columnwidth]{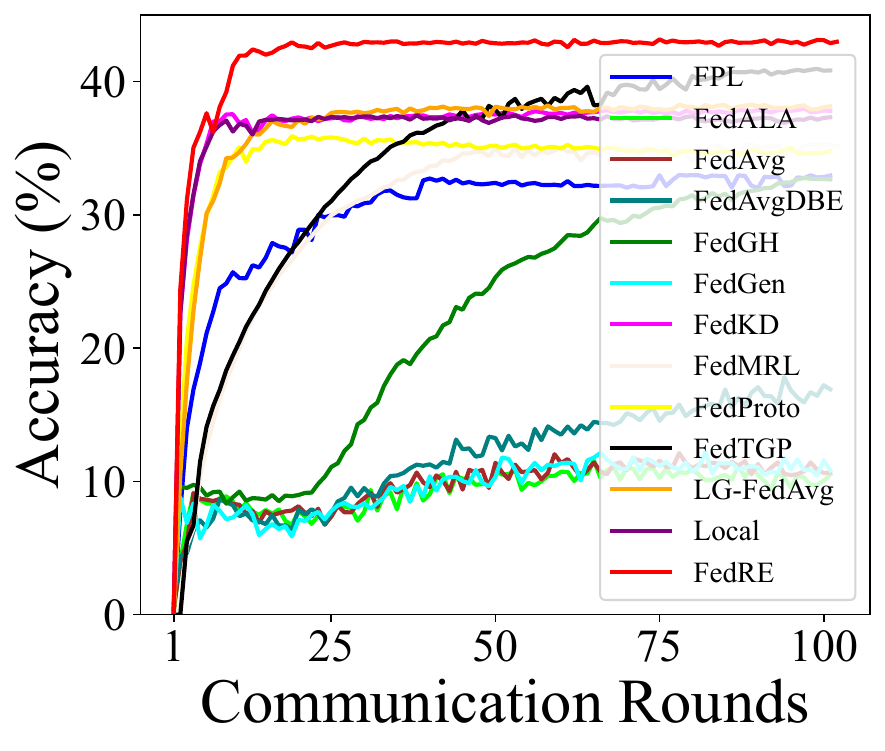}}
\caption{Accuracy (\%) comparison between distinct communication rounds on the TinyImageNet dataset in the model-homogeneous FL setting in both the PRA and PAT settings.}
\label{appdix:Fig:accu}
\end{figure}

Model-homogeneous FL can be regarded as a special case of model-heterogeneous FL, where all clients adopt the same model architecture. In our experiments, we adopt a four-layer CNN for CIFAR-10 and CIFAR-100, and ResNet-18 for TinyImageNet across all methods. \Cref{tab:model-homogeneous} presents the results, where FedRE achieves the best performance across all datasets. Specifically, FedRE's average accuracy is 63.21\%, outperforming the second-best method, \textit{i.e.}, LG-FedAvg, by 2.58\%. Figure~\ref{appdix:Fig:accu} provides convergence comparisons on TinyImageNet, where FedRE exhibits a rapid initial improvement followed by gradual stabilization, indicating stable convergence behavior throughout training. Those results suggest that FedRE remains effective in model-homogeneous FL.

\begin{table}[htbp]
\centering
\caption{Accuracy (\%) comparison on three datasets in the model-homogeneous setting. In each column, the best results are \textbf{bolded}, and the second-best results are \underline{underlined}.}
\label{tab:model-homogeneous} 
\begin{tabular}{@{}c@{\hskip 5pt}c@{\hskip 5pt}c@{\hskip 5pt}c@{\hskip 5pt}c@{\hskip 5pt}c@{\hskip 5pt}c@{\hskip 5pt}c@{}}
\toprule
\multirow{2}{*}{\textbf{Method}} & \multicolumn{3}{c}{\textbf{PRA}} & \multicolumn{3}{c}{\textbf{PAT}} & \multirow{2}{*}{\textbf{Average}} \\
\cmidrule(lr){2-4} \cmidrule(lr){5-7}
& \textbf{CIFAR-10} & \textbf{CIFAR-100} & \textbf{TinyImageNet} & \textbf{CIFAR-10} & \textbf{CIFAR-100} & \textbf{TinyImageNet} & \\ 
\midrule
LG-FedAvg \cite{liang2020think} & \underline{86.92 $\pm$ 0.25} & 49.82 $\pm$ 0.39 & \underline{32.00 $\pm$ 0.13} & 90.59 $\pm$ 0.17 & 66.00 $\pm$ 0.27 & 38.43 $\pm$ 0.23 & \underline{60.63} \\
FedAvg \cite{mcmahan2017communication} & 55.21 $\pm$ 0.12 & 30.37 $\pm$ 0.02 & 13.66 $\pm$ 0.41 & 52.70 $\pm$ 0.11 & 24.89 $\pm$ 0.20 & 9.98 $\pm$ 0.48 & 31.14 \\
FedALA \cite{zhang2023fedala} & 55.02 $\pm$ 0.14 & 29.89 $\pm$ 0.22 & 13.63 $\pm$ 0.10 & 52.83 $\pm$ 0.19 & 24.91 $\pm$ 0.15 & 10.65 $\pm$ 0.15 & 31.16 \\
FedGH \cite{yi2023fedgh} & 86.02 $\pm$ 0.17 & 48.59 $\pm$ 0.60 & 28.64 $\pm$ 0.26 & 90.46 $\pm$ 0.22 & 65.14 $\pm$ 0.26 & 32.40 $\pm$ 0.19 & 58.54 \\
FedKD \cite{Wu2022Communication} & 86.23 $\pm$ 0.12 & \underline{51.91 $\pm$ 0.28} & 29.47 $\pm$ 0.31 & 90.01 $\pm$ 0.09 & 67.23 $\pm$ 0.38 & 35.34 $\pm$ 0.33 & 60.03 \\
FedAvgDBE \cite{Zhang2023Eliminating} & 78.10 $\pm$ 0.20 & 35.23 $\pm$ 0.24 & 16.92 $\pm$ 0.52 & 82.27 $\pm$ 0.45 & 35.21 $\pm$ 0.27 & 16.80 $\pm$ 0.23 & 44.09 \\
FedGen \cite{Zhu2021Data} & 55.21 $\pm$ 0.14 & 29.90 $\pm$ 0.17 & 13.76 $\pm$ 0.23 & 52.37 $\pm$ 0.22 & 24.82 $\pm$ 0.38 & 10.67 $\pm$ 0.54 & 31.12 \\
FedProto \cite{tan2022fedproto} & 85.63 $\pm$ 0.22 & 50.52 $\pm$ 0.19 & 28.67 $\pm$ 0.17 & \underline{91.04 $\pm$ 0.16} & \underline{69.28 $\pm$ 0.07} & 34.75 $\pm$ 0.49 & 59.98 \\
FPL \cite{huang2023rethinking} & 83.60 $\pm$ 0.03 & 49.10 $\pm$ 0.21 & 26.87 $\pm$ 0.09 & 90.59 $\pm$ 0.06 & 67.31 $\pm$ 0.03 & 32.95 $\pm$ 0.23&  58.40  \\
FedMRL \cite{yi2024federated} & 82.55 $\pm$ 0.31 & 48.41 $\pm$ 0.09 & 26.78 $\pm$ 0.05 & 89.02 $\pm$ 0.23 & 65.97 $\pm$ 0.21 & 35.22 $\pm$ 0.05 & 57.99 \\
FedTGP \cite{zhang2024fedtgp} & 85.59 $\pm$ 0.12 & 47.05 $\pm$ 0.17 & 30.89 $\pm$ 0.21 & 90.49 $\pm$ 0.03 & 67.47 $\pm$ 0.07 & \underline{40.88 $\pm$ 0.20} & 60.40 \\
Local & 86.33 $\pm$ 0.11 & 49.88 $\pm$ 0.40 & 31.44 $\pm$ 0.15 & 90.54 $\pm$ 0.12 & 66.57 $\pm$ 0.17 & 37.46 $\pm$ 0.27 & 60.37 \\
FedRE & \textbf{86.99 $\pm$ 0.01} & \textbf{52.12 $\pm$ 0.04} & \textbf{36.12 $\pm$ 0.21} & \textbf{91.06 $\pm$ 0.01} & \textbf{70.52 $\pm$ 0.17} & \textbf{42.45 $\pm$ 0.17} & \textbf{63.21} \\
\bottomrule
\end{tabular}%
\vspace{-2ex}
\end{table}

\subsection{Impact of a Small Number of Clients}
\label{appendix:SmallClients}

We conduct experiments on CIFAR-10 and CIFAR-100 to evaluate the impact of a small number of clients ($K \in \{2, 4, 10\}$) in the model-homogeneous setting, using a four-layer CNN. \Cref{tab:smallClients} reports the results, from which we draw several observations. 
(1) Both FedGH and FedRE experience performance degradation as $K$ decreases, due to reduced diversity in the prototypes or entangled representations used to train the global classifier. 
(2) Despite this trend, FedProto, FedGH, and FedRE consistently outperform FedAvg, demonstrating their effectiveness even with fewer clients. 
(3) FedRE achieves the best performance across all settings. Specifically, with only 2 clients, it reaches 41.57\% on CIFAR-100, outperforming FedAvg by 8.44\%, FedProto by 3.70\%, and FedGH by 1.96\%. 
Those results indicate that FedRE remains effective even with a small number of clients.

\begin{table}[htbp]
\centering
\caption{Accuracy (\%)  comparison on CIFAR-10 and CIFAR-100 with different numbers of clients. In each column, the best results are \textbf{bolded}, and the second-best results are \underline{underlined}.}
\label{tab:smallClients}
\begin{tabular}{l|c|c|c|c|c|c}
\toprule
\multirow{2}{*}{Method} & \multicolumn{3}{c|}{CIFAR-10} & \multicolumn{3}{c}{CIFAR-100} \\
\cline{2-7}
 & 2 & 4 & 10 & 2 & 4 & 10 \\
\midrule
FedAvg \cite{mcmahan2017communication} & 64.99 & 62.26 & 55.21 & 33.13 & 30.88 & 30.37 \\
FedProto \cite{tan2022fedproto} & 77.09 & \underline{84.69} & 85.63 & 37.87 & 44.85 & \underline{50.52} \\
FedGH \cite{yi2023fedgh} & \underline{78.80} & 84.37 & \underline{86.02} & \underline{39.61} & \underline{46.40} & 48.59 \\
FedRE & \textbf{79.11} & \textbf{85.09} & \textbf{86.99} & \textbf{41.57} & \textbf{47.86} & \textbf{52.12} \\
\bottomrule
\end{tabular}
\end{table}

\subsection{Privacy Protection Evaluation}
\label{appendix:PrivacyProtectionEvaluation}

\begin{figure}[htbp]
    \centering    
    \subfloat[Original Images]{        
        \includegraphics[width=1.65cm,height=1.65cm]{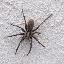}
        \includegraphics[width=1.65cm,height=1.65cm]{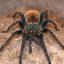}
        \includegraphics[width=1.65cm,height=1.65cm]{Image/preimage/input/class0_0.JPEG}
        \includegraphics[width=1.65cm,height=1.65cm]{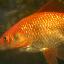}
        \includegraphics[width=1.65cm,height=1.65cm]{Image/preimage/input/class4_33.JPEG}
        \includegraphics[width=1.65cm,height=1.65cm]{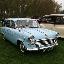}
        \includegraphics[width=1.65cm,height=1.65cm]{Image/preimage/input/class2_5.JPEG}
        \includegraphics[width=1.65cm,height=1.65cm]{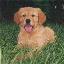}
        \includegraphics[width=1.65cm,height=1.65cm]{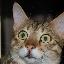}
        \includegraphics[width=1.65cm,height=1.65cm]{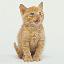}
        }
    \\
    \vspace{-1ex}
    \centering
    \subfloat[Reconstruction Images from Representations]{
    \includegraphics[width=1.65cm,height=1.65cm]{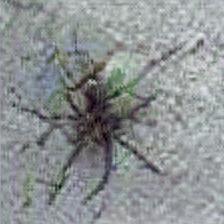}        \includegraphics[width=1.65cm,height=1.65cm]{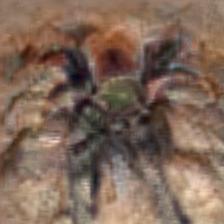}
    \includegraphics[width=1.65cm,height=1.65cm]{Image/preimage/rep/class0_0.JPEG}
        \includegraphics[width=1.65cm,height=1.65cm]{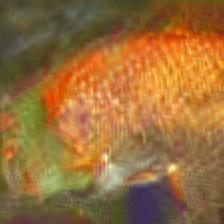}
        \includegraphics[width=1.65cm,height=1.65cm]{Image/preimage/rep/class4_33.JPEG}
        \includegraphics[width=1.65cm,height=1.65cm]{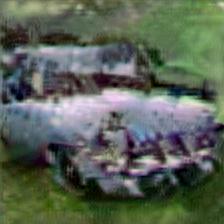}
        \includegraphics[width=1.65cm,height=1.65cm]{Image/preimage/rep/class2_5.JPEG}
        \includegraphics[width=1.65cm,height=1.65cm]{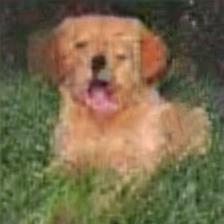}
        \includegraphics[width=1.65cm,height=1.65cm]{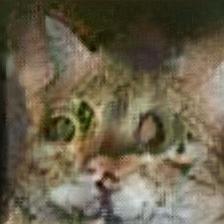}
        \includegraphics[width=1.65cm,height=1.65cm]{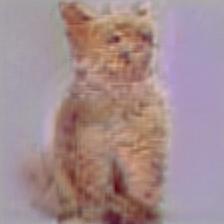}
        }
    \\
    \vspace{-1ex}
    \centering
    \subfloat[Reconstruction Images from Prototypes]{
        \includegraphics[width=1.65cm,height=1.65cm]{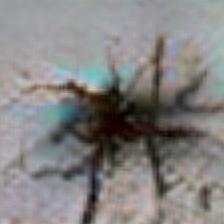}
        \includegraphics[width=1.65cm,height=1.65cm]{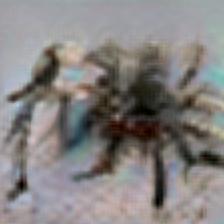}
        \includegraphics[width=1.65cm,height=1.65cm]{Image/preimage/proto/class0_0.JPEG}
        \includegraphics[width=1.65cm,height=1.65cm]{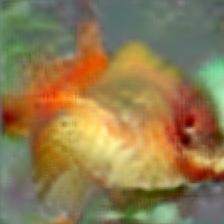}
        \includegraphics[width=1.65cm,height=1.65cm]{Image/preimage/proto/class4_30.JPEG}
        \includegraphics[width=1.65cm,height=1.65cm]{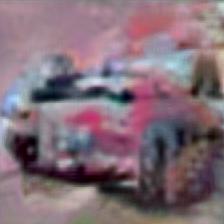}
        \includegraphics[width=1.65cm,height=1.65cm]{Image/preimage/proto/class2_65.JPEG}
        \includegraphics[width=1.65cm,height=1.65cm]{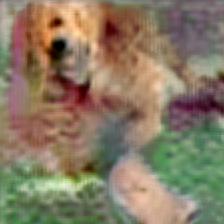}
        \includegraphics[width=1.65cm,height=1.65cm]{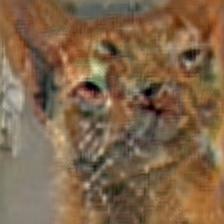}
        \includegraphics[width=1.65cm,height=1.65cm]{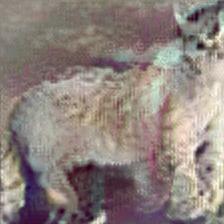}
        }
    \\
    \vspace{-1ex}
    \centering
    \subfloat[Reconstruction Images from Entangled Representations]{
    \includegraphics[width=1.65cm,height=1.65cm]{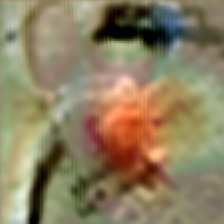}       \includegraphics[width=1.65cm,height=1.65cm]{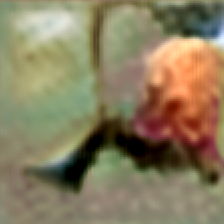}
    \includegraphics[width=1.65cm,height=1.65cm]{Image/preimage/fuse/0.png}
    \includegraphics[width=1.65cm,height=1.65cm]{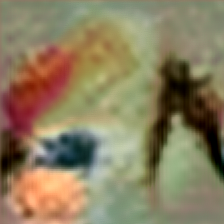}       \includegraphics[width=1.65cm,height=1.65cm]{Image/preimage/fuse/2.png}
    \includegraphics[width=1.65cm,height=1.65cm]{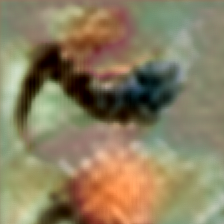}
    \includegraphics[width=1.65cm,height=1.65cm]{Image/preimage/fuse/4.png}
    \includegraphics[width=1.65cm,height=1.65cm]{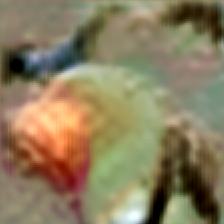}
    \includegraphics[width=1.65cm,height=1.65cm]{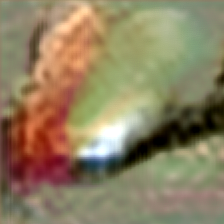}
    \includegraphics[width=1.65cm,height=1.65cm]{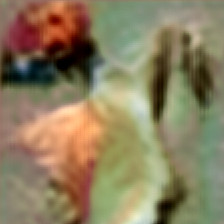}
    }
\caption{Comparison of privacy protection in restructuring results from representations, prototypes, and entangled representations on the TinyImageNet dataset.}
\label{Fig:privacy-ap}
\end{figure}

\cref{Fig:privacy-ap} presents additional reconstructed results on sample images from the TinyImageNet dataset. As can be seen, images reconstructed from the entangled representations contain less discernible content or category information than those reconstructed from vanilla representations or prototypes, suggesting that FedRE offers a good level of privacy protection against representation inversion attacks.

\subsection{Communication Overhead Evaluation}
\label{appendix:CommunicationOverheadEvaluation}

\Cref{appendix:tab:CO} lists the communication overhead results on the CIFAR-10, CIFAR-100, and TinyImageNet datasets under both model-heterogeneous (Model-hete) and model-homogeneous (Model-homo) scenarios in the PRA setting. 
As can be seen, FedRE generally achieves lower communication overhead than the baselines, which suggests its potential effectiveness in reducing communication overhead.

\begin{table}[htbp]
    \centering
    \vspace{-1ex}
    \caption{Communication overhead (\# Scalars $\times 10^3$) comparison on three datasets. In each column, the best results are \textbf{bolded}, and the second-best results are \underline{underlined}.}
    \label{appendix:tab:CO} 
    \resizebox{\textwidth}{!}{%
    \begin{tabular}{@{}c*{12}{@{\hskip 0.2pt}c}@{}}
    \toprule
    \multirow{3}{*}{\textbf{Method}} & \multicolumn{4}{c}{\textbf{CIFAR-10}} & \multicolumn{4}{c}{\textbf{CIFAR-100}} & \multicolumn{4}{c}{\textbf{TinyImageNet}} \\
    \cmidrule(lr){2-5} \cmidrule(lr){6-9} \cmidrule(lr){10-13}
    & \multicolumn{2}{c}{\textbf{Model-homo}} & \multicolumn{2}{c}{\textbf{Model-hete}} & \multicolumn{2}{c}{\textbf{Model-homo}} & \multicolumn{2}{c}{\textbf{Model-hete}} & \multicolumn{2}{c}{\textbf{Model-homo}} & \multicolumn{2}{c}{\textbf{Model-hete}} \\
    \cmidrule(lr){2-3} \cmidrule(lr){4-5} \cmidrule(lr){6-7} \cmidrule(lr){8-9} \cmidrule(lr){10-11} \cmidrule(lr){12-13}
    & \textbf{Upload} & \textbf{Broadcast} & \textbf{Upload} & \textbf{Broadcast} & \textbf{Upload} &\textbf{ Broadcast} & \textbf{Upload} & \textbf{Broadcast} &\textbf{ Upload} &\textbf{ Broadcast} & \textbf{Upload} & \textbf{Broadcast} \\
    \midrule
    LG-FedAvg  & 51.30 & \underline{51.30} & 51.30 & \underline{51.30} & 513.00 & \underline{513.00} & 513.00 & \underline{513.00} & 4098.00 & \underline{4098.00} & 4098.00 & \underline{4098.00}  \\
    FedGH  & \underline{31.23} & \textbf{51.20} & \underline{31.23} & \textbf{51.20} & \underline{257.02} & \textbf{512.00} & \underline{257.02} & \textbf{512.00} & \underline{1918.98} & \textbf{4096.00} & \underline{1918.98} & \textbf{4096.00} \\
    FedKD & 3374.28 & 3374.28 & 3353.68 & 3353.68 & 4234.28 & 4234.28 & 3524.67 & 3524.67 & 90503.00 & 90503.00 & 57544.97 & 57544.97\\
    FedGen & 8785.38 & 8785.38 & 51.30 & \underline{51.30} & 9247.08 & 9247.08 & 513.00 & \underline{513.00} & 239178.32 & 239178.32 & 4098.00 & \underline{4098.00}  \\
    FedProto  & \underline{31.23} & \textbf{51.20} & \underline{31.23} & \textbf{51.20} & \underline{257.02} & \textbf{512.00} & \underline{257.02} & \textbf{512.00} & \underline{1918.98} & \textbf{4096.00} & \underline{1918.98} & \textbf{4096.00}  \\
    FPL  & \underline{31.23} & 87.04 & \underline{31.23} & 112.64 & \underline{257.02} & 916.48 & \underline{257.02} & 1182.72 & \underline{1918.98} & 9768.96 & \underline{1918.98} & 10567.68   \\
    FedMRL   & 8746.98 & 8746.98& 8746.98& 8746.98& 8863.08& 8863.08& 8863.08& 8863.08& 56178.00 & 56178.00 & 56178.00 & 56178.00  \\
    FedTGP & \underline{31.23} & \textbf{51.20} & \underline{31.23} & \textbf{51.20} & \underline{257.02} & \textbf{512.00} & \underline{257.02} & \textbf{512.00} & \underline{1918.98} & \textbf{4096.00} & \underline{1918.98} & \textbf{4096.00}  \\
    \midrule
    FedAvg  & 8785.38 & 8785.38 & - & - & 9247.08 & 9247.08 & - & - & 239178.32 & 239178.32 & - & -  \\
    FedALA  & 8785.38 & 8785.38 & - & - & 9247.08 & 9247.08 & - & - & 239178.32 & 239178.32 & - & - \\
    FedAvgDBE  & 8785.38 & 8785.38 & - & - & 9247.08 & 9247.08 & - & - & 239178.32 & 239178.32 & - & -  \\
    \midrule
    FedRE & \textbf{5.12} & \underline{51.30} & \textbf{5.12} & \underline{51.30} & \textbf{5.12} & \underline{513.00} & \textbf{5.12} & \underline{513.00} & \textbf{20.48} & \underline{4098.00} & \textbf{20.48} & \underline{4098.00}  \\
    \bottomrule
    \end{tabular}%
    }
\vspace{-2ex}
\end{table}

\subsection{Statistical Heterogeneity Analysis}
\label{appendix:statisticalheterogeneityanalysis}

\begin{figure}[H]
    \centering
    \subfloat[\small CIFAR-10 ($\alpha$ = 0.05)]{
        {\includegraphics[width=4.1cm,height=3.3cm]{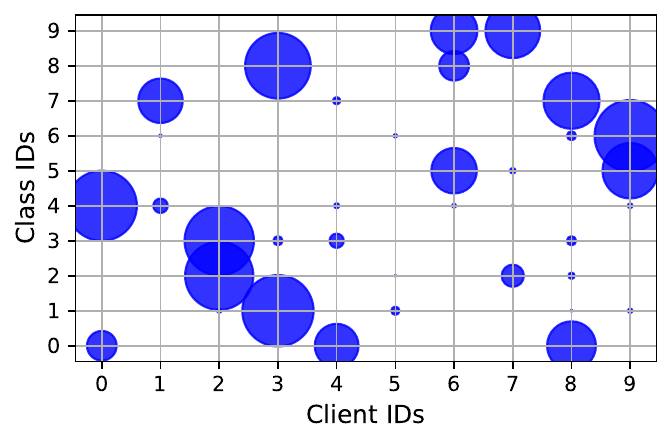}}
    }\subfloat[\small CIFAR-10 ($\alpha$ = 0.1)]{
        {\includegraphics[width=4.1cm,height=3.3cm]{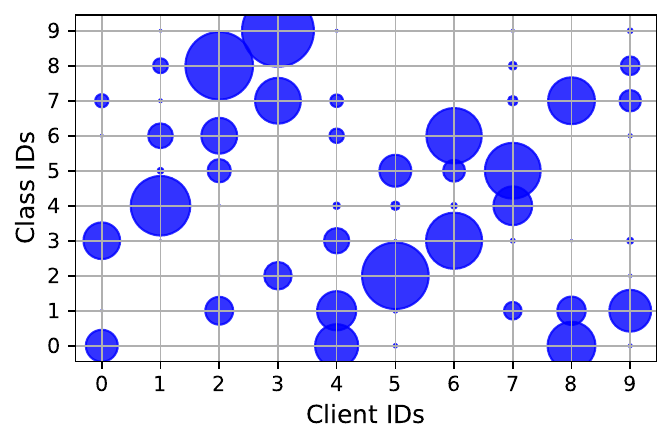}}
    }
    \subfloat[\small CIFAR-10 ($\alpha$ = 1)]{
        {\includegraphics[width=4.1cm,height=3.3cm]{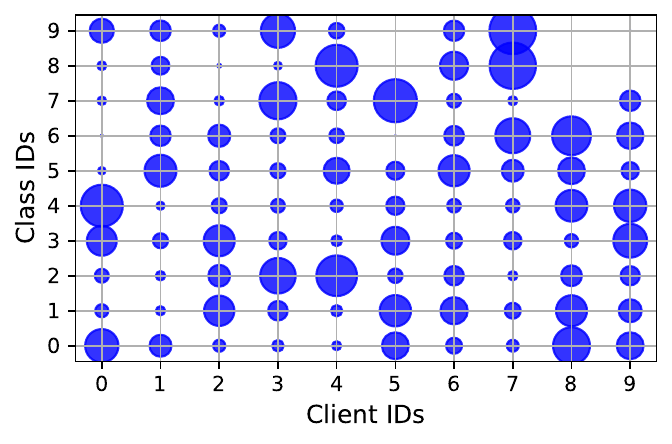}}
    }
    \subfloat[\small CIFAR-10 ($\alpha$ = 10)]{
        {\includegraphics[width=4.1cm,height=3.3cm]{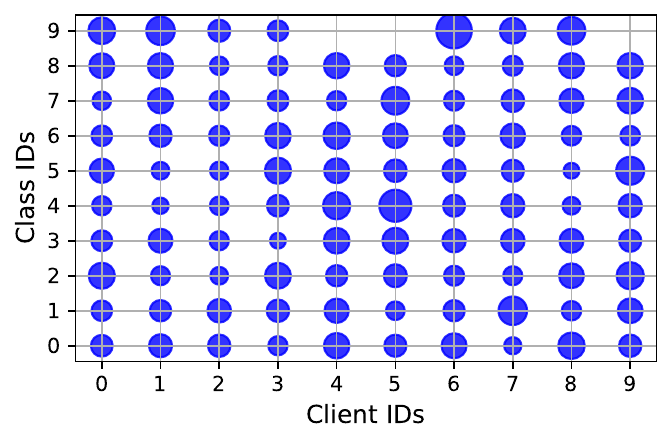}}
    }\\
\subfloat[\small CIFAR-100 ($\alpha$ = 0.05)]{
        {\includegraphics[width=0.23\columnwidth]{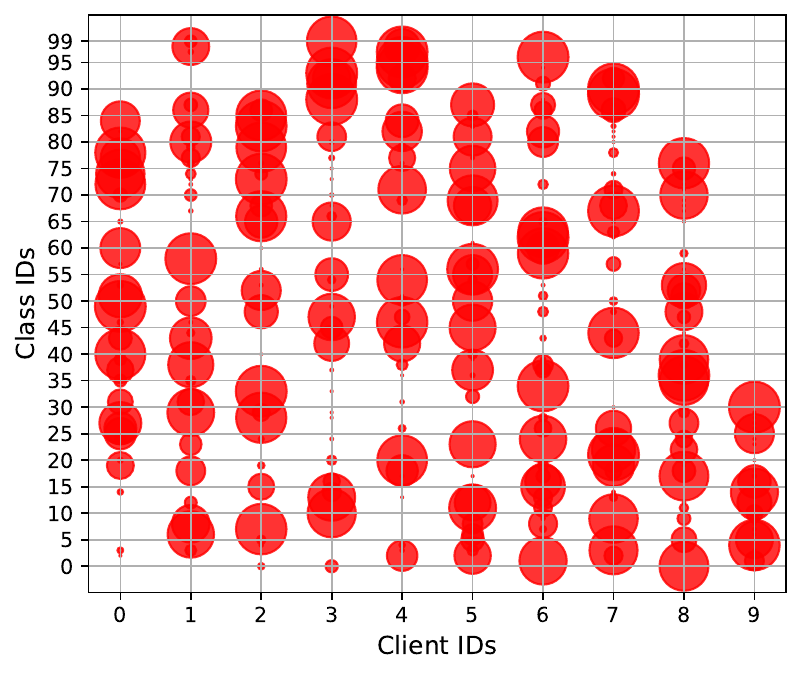}}
    }\subfloat[\small CIFAR-100 ($\alpha$ = 0.1)]{
        {\includegraphics[width=0.23\columnwidth]{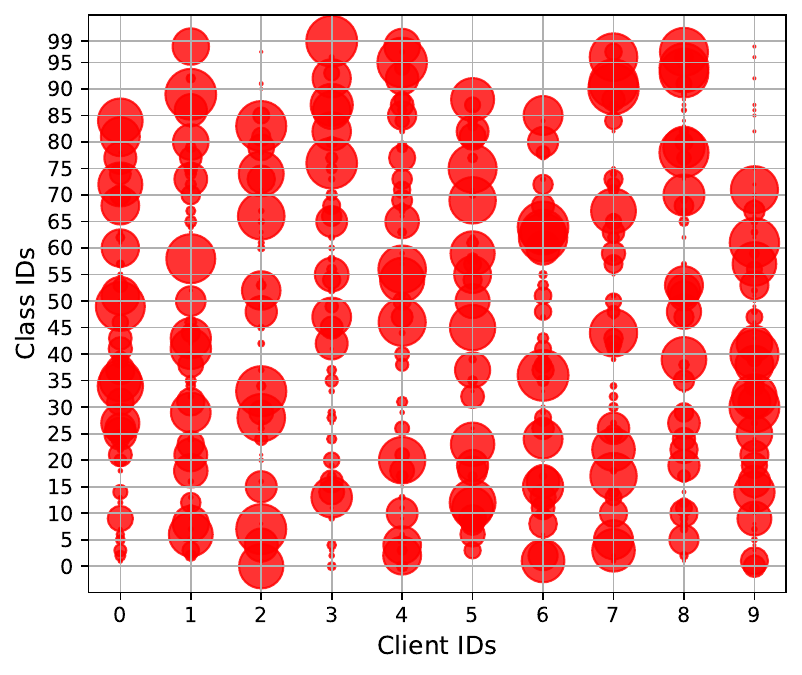}}
    }
    \subfloat[\small CIFAR-100 ($\alpha$ = 1)]{
        {\includegraphics[width=0.23\columnwidth]{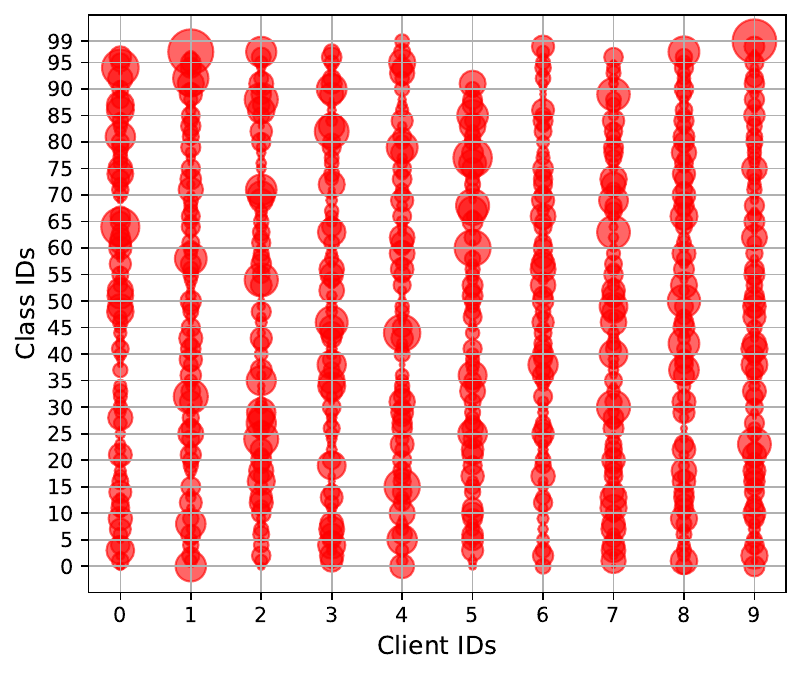}}
    }
    \subfloat[\small CIFAR-100 ($\alpha$ = 10)]{
        {\includegraphics[width=0.23\columnwidth]{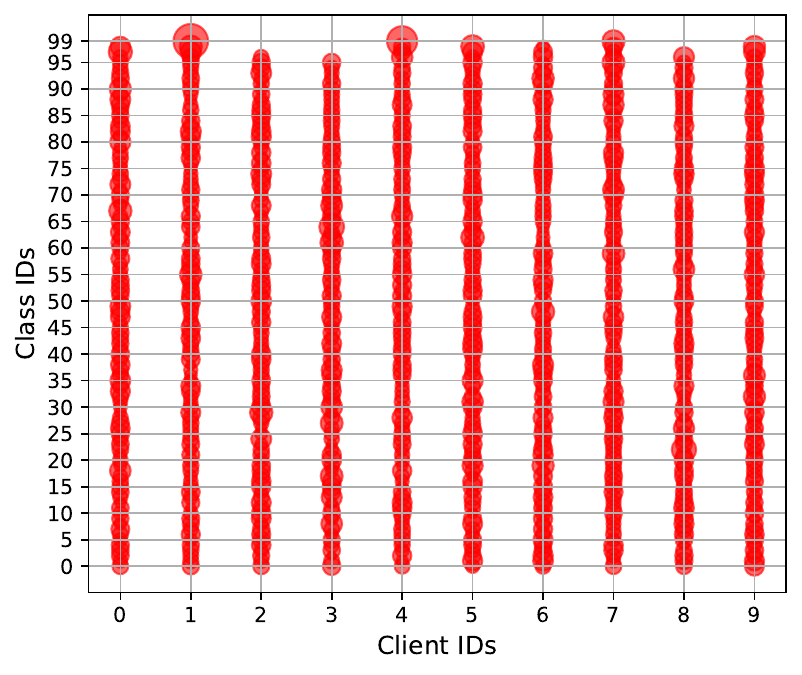}}
    }    
\caption{The sample distributions for all clients on the CIFAR-10 and CIFAR-100 datasets under the PRA settings with varying parameters $\alpha$. The size of each circle indicates the number of samples.}
\label{appendix:fig:datavisualPRA}
\vspace{-1ex}
\end{figure}

\begin{figure}[htbp]
    \centering
    \vspace{-1ex}
    \subfloat[\small CIFAR-10 (10 Clients)]{
         {\includegraphics[width=4.1cm,height=3.3cm]{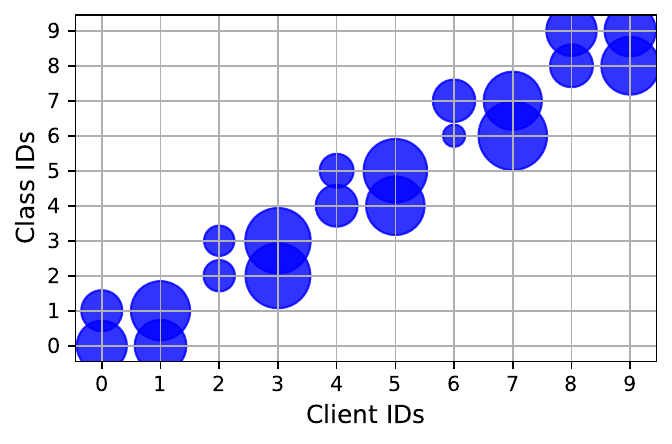}}
    }
    \subfloat[\small CIFAR-10 (25 Clients)]{
         {\includegraphics[width=4.1cm,height=3.3cm]{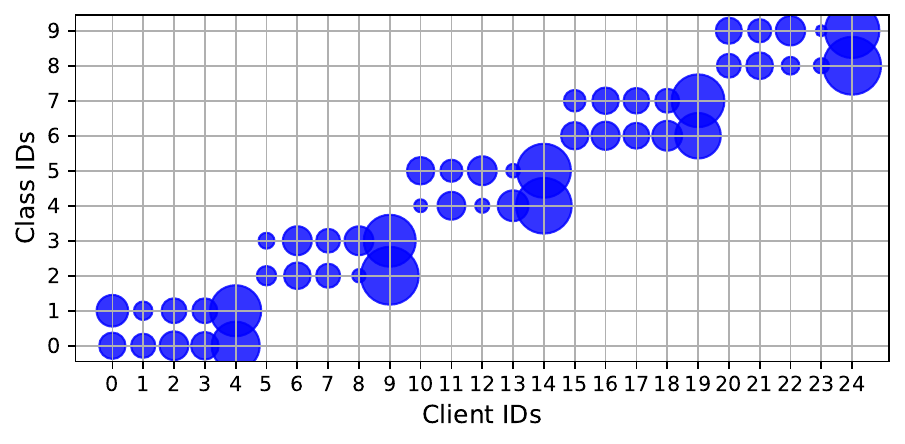}}
    }
    \subfloat[\small CIFAR-100 (10 Clients)]{
         {\includegraphics[width=4.1cm,height=3.3cm]{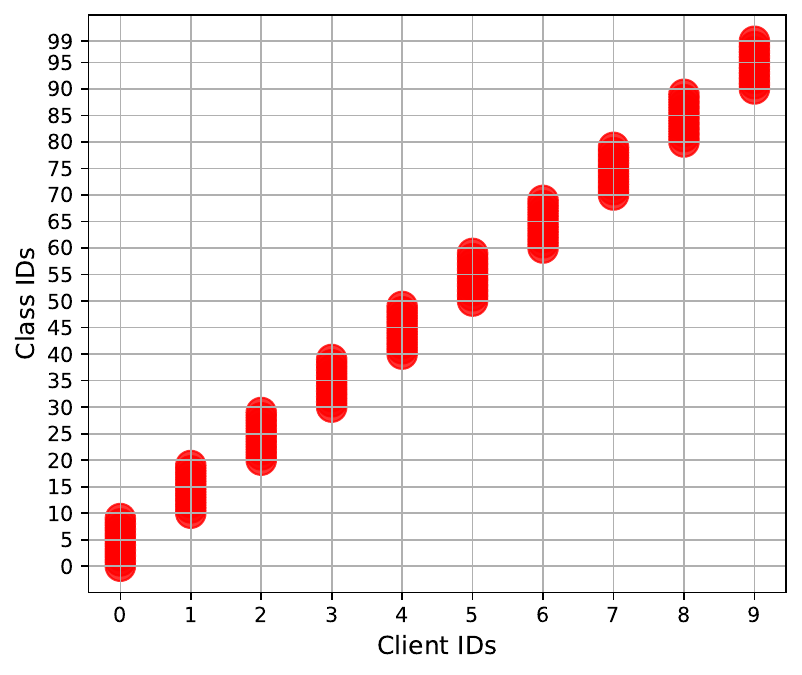}}
    }
    \subfloat[\small CIFAR-100 (25 Clients)]{
         {\includegraphics[width=4.1cm,height=3.3cm]{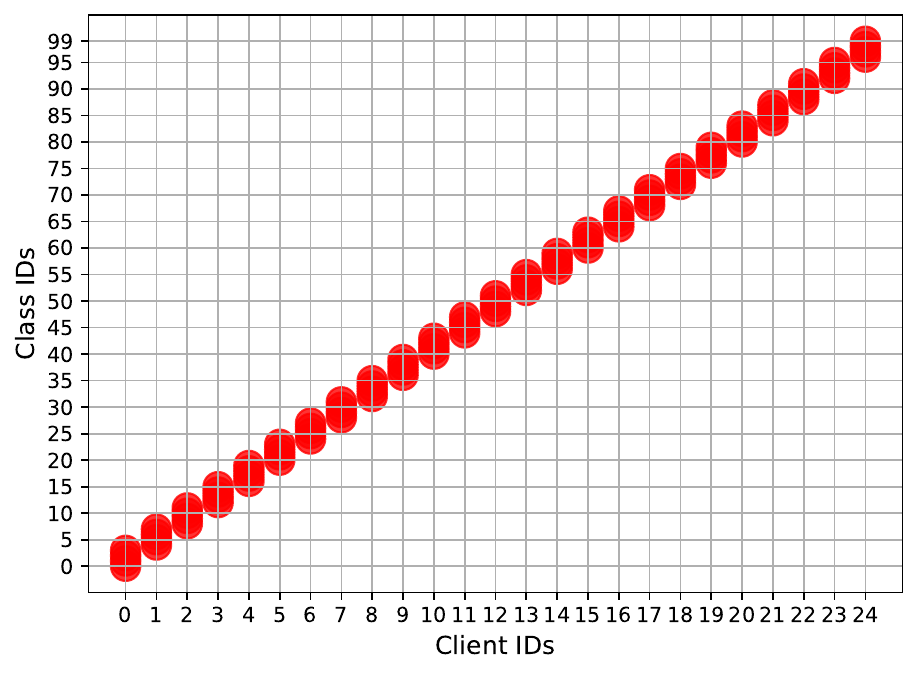}}
    }
\caption{The sample distributions for all clients on the CIFAR-10 and CIFAR-100 datasets under the PAT settings with varying client numbers. The size of each circle indicates the number of samples.}
\label{appendix:fig:datavisualPAT}
\vspace{-1ex}
\end{figure}

\begin{figure}[htbp]
    \centering
    \vspace{-1ex}
    \subfloat[\small CIFAR-10 with PRA]{
        \includegraphics[width=0.245\columnwidth]{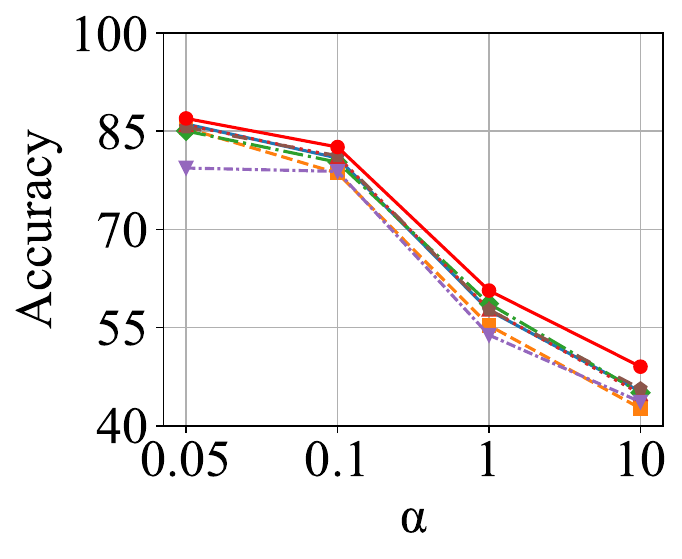}}
    \subfloat[\small CIFAR-100 with PRA]
    {\includegraphics[width=0.235\columnwidth]{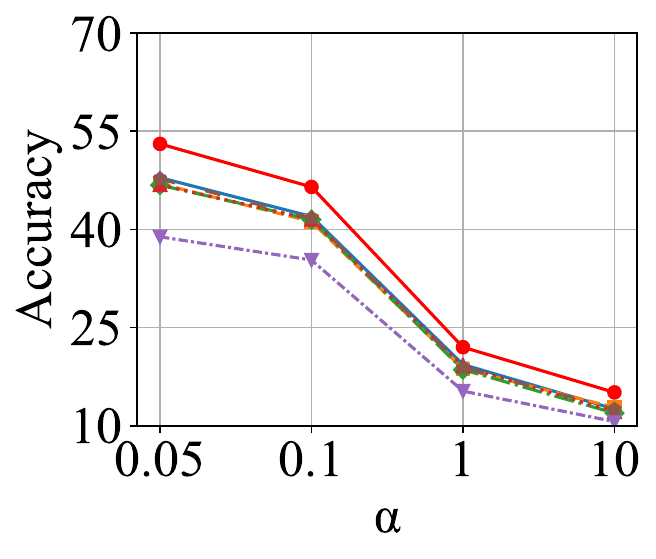}}
    \subfloat[\small CIFAR-10 with PAT]{
        \includegraphics[width=0.25\columnwidth]{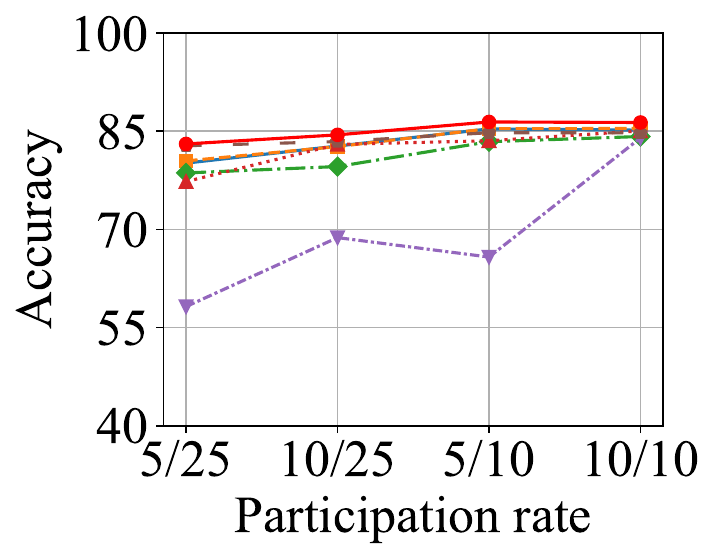}}
\subfloat[\small CIFAR-100 with PAT]{\includegraphics[width=0.238\columnwidth]{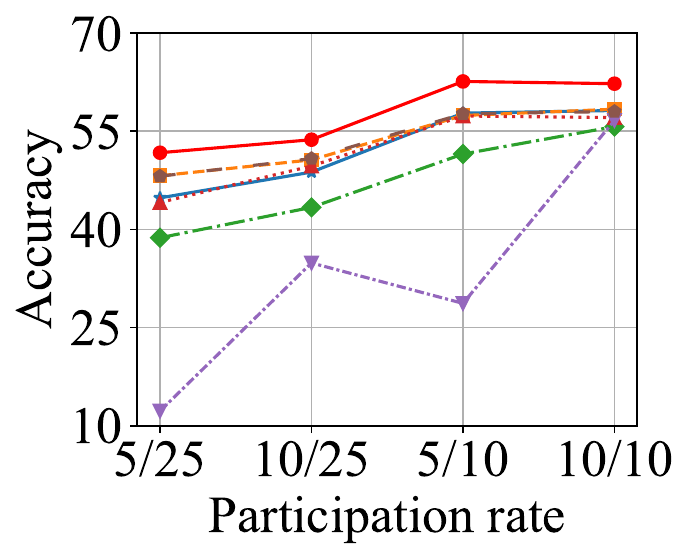}}\\
    \subfloat{
        \includegraphics[width=0.95\textwidth]{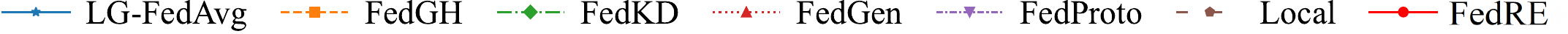}}
\caption{Accuracy (\%) comparison between distinct statistic-heterogeneous scenarios on the CIFAR-10 and CIFAR-100 datasets.}
\label{fig:varystatitich}
\end{figure}

To further evaluate the effectiveness of FedRE under different levels of statistical heterogeneity, we adjust the Dirichlet distribution parameter $\alpha$ (\textit{i.e.}, 0.05, 0.1, 1, 10) in the PRA setting and the client participation rate (\textit{i.e.}, 5/25, 10/25 for 25 clients, 5/10, 10/10 for 10 clients) in the PAT setting, respectively, to control the degree of sample skewness. The resulting sample distributions are visualized in Figures~\ref{appendix:fig:datavisualPRA}-\ref{appendix:fig:datavisualPAT}. 
The results on the CIFAR-10 and CIFAR-100 datasets, under the model-heterogeneous setting, are shown in \cref{fig:varystatitich}. 
As shown, FedRE achieves competitive accuracy across different levels of statistical heterogeneity, suggesting it remains effective under varying degrees of data heterogeneity.


\vfill

\end{document}